\documentclass[10pt,twocolumn,letterpaper]{article}

\usepackage{graphicx}	
\usepackage{amsmath}	
\usepackage{amssymb}	
\usepackage{booktabs}
\usepackage{times}
\usepackage{microtype}
\usepackage{epsfig}
\usepackage{caption}
\usepackage{float}
\usepackage{placeins}
\usepackage{color, colortbl}
\usepackage{stfloats}
\usepackage{tabularx}
\usepackage{xstring}
\usepackage{multirow}
\usepackage{xspace}
\usepackage{subcaption}
\usepackage{utfsym}
\usepackage[hang,flushmargin]{footmisc}
\usepackage{algorithm}
\usepackage{listings}

% ...

\newcommand{\nbf}[1]{{\noindent \textbf{#1}}}

\newcommand{\R}[1]{{%
    \textbf{%
        \ifstrequal{#1}{1}{\textcolor{red}{R#1}}{%
        \ifstrequal{#1}{2}{\textcolor{blue}{R#1}}{%
        \ifstrequal{#1}{3}{\textcolor{magenta}{R#1}}{%
        \ifstrequal{#1}{4}{\textcolor{teal}{R#1}}{%
                           \textcolor{cyan}{R#1}%
        }}}}%
    }%
}}

\newcommand{\tablestyle}[2]{\setlength{\tabcolsep}{#1}\renewcommand{\arraystretch}{#2}\centering\footnotesize}

\definecolor{darkgray}{HTML}{757575}

\usepackage[pagenumbers]{iccv} 

\DeclareUnicodeCharacter{FF0C}{,}

\definecolor{iccvblue}{rgb}{0.21,0.49,0.74}
\usepackage[pagebackref,breaklinks,colorlinks,allcolors=iccvblue]{hyperref}

\def\paperID{4125} 
\def\confName{ICCV}
\def\confYear{2025}

\title{SP$^2$T: Sparse Proxy Attention for Dual-stream Point Transformer}

\def\authorBlock{
    Jiaxu Wan$^{1,}$\thanks{Co-first author}, 
    Hong Zhang$^{1,}$\footnotemark[1],
    Ziqi He$^{2,}$\footnotemark[1],
    Yangyan Deng$^{3}$,
    Qishu Wang$^{2}$, 
    Ding Yuan$^{1}$, 
    Yifan Yang$^{1,}$\thanks{Corresponding author}
    \\
    $^{1}$School of Aerospace, BUAA \qquad 
    $^{2}$School of Artificial Intelligence, BUAA\\
    $^{3}$School of Electronics and Information Engineering, BUAA\\
    \href{https://github.com/WallelWan/SP2T}{https://github.com/WallelWan/SP2T}
}

\author{\authorBlock}

\begin{document}
\maketitle

\begin{abstract}
Point transformers have demonstrated remarkable progress in 3D understanding through expanded receptive fields (RF), but further expanding the RF leads to dilution in group attention and decreases detailed feature extraction capability. Proxy, which serves as abstract representations for simplifying feature maps, enables global RF. However, existing proxy-based approaches face critical limitations: Global proxies incur quadratic complexity for large-scale point clouds and suffer positional ambiguity, while local proxy alternatives struggle with 1) Unreliable sampling from the geometrically diverse point cloud, 2) Inefficient proxy interaction computation, and 3) Imbalanced local-global information fusion;
To address these challenges, we propose Sparse Proxy Point Transformer (SP$^{2}$T) -- a local proxy-based dual-stream point transformer with three key innovations: \textit{First}, for reliable sampling, spatial-wise proxy sampling with vertex-based associations enables robust sampling on geometrically diverse point clouds. \textit{Second}, for efficient proxy interaction, sparse proxy attention with a table-based relative bias effectively achieves the interaction with efficient map-reduce computation. \textit{Third}, for local-global information fusion, our dual-stream architecture maintains local-global balance through parallel branches.
Comprehensive experiments reveal that SP$^{2}$T sets state-of-the-art results with acceptable latency on indoor and outdoor 3D comprehension benchmarks, demonstrating marked improvement (+3.8\% mIoU vs. SPoTr@S3DIS, +22.9\% mIoU vs. PointASNL@Sem.KITTI) compared to other proxy-based point cloud methods.
\end{abstract}

\section{Introduction}


The point transformer~\cite{zhao2021point, wu2022point, park2022fast, wu2024point} has recently emerged as a key research area in 3D perception~\cite{wang2022detr3d, hao2024mapdistill, li2024mlp} and other fields~\cite{zhang2024p2ftrack}, mainly using group attention~\cite{liu2021swin, liu2022swin} to extract features. Although various point transformer methods~\cite{wu2022point, wu2024point} have recently advanced in expanding receptive fields (RF), they remain constrained by the large scale of points. However, expanding RF within PTv3 can lead to attention dilution~\cite{park2022vision}, where a larger group size diverts the model's focus, impairing its ability to capture precise features (77.3 mIOU with 1024 points vs. 77.1 mIOU with 4096 points@ScanNetv2~\cite{dai2017scannet}). Furthermore, this expansion is also computationally demanding. Consequently, the broadening of the RF must be achieved differently.



\begin{figure}[t]
    \centering
    \includegraphics[width=1.0\linewidth]{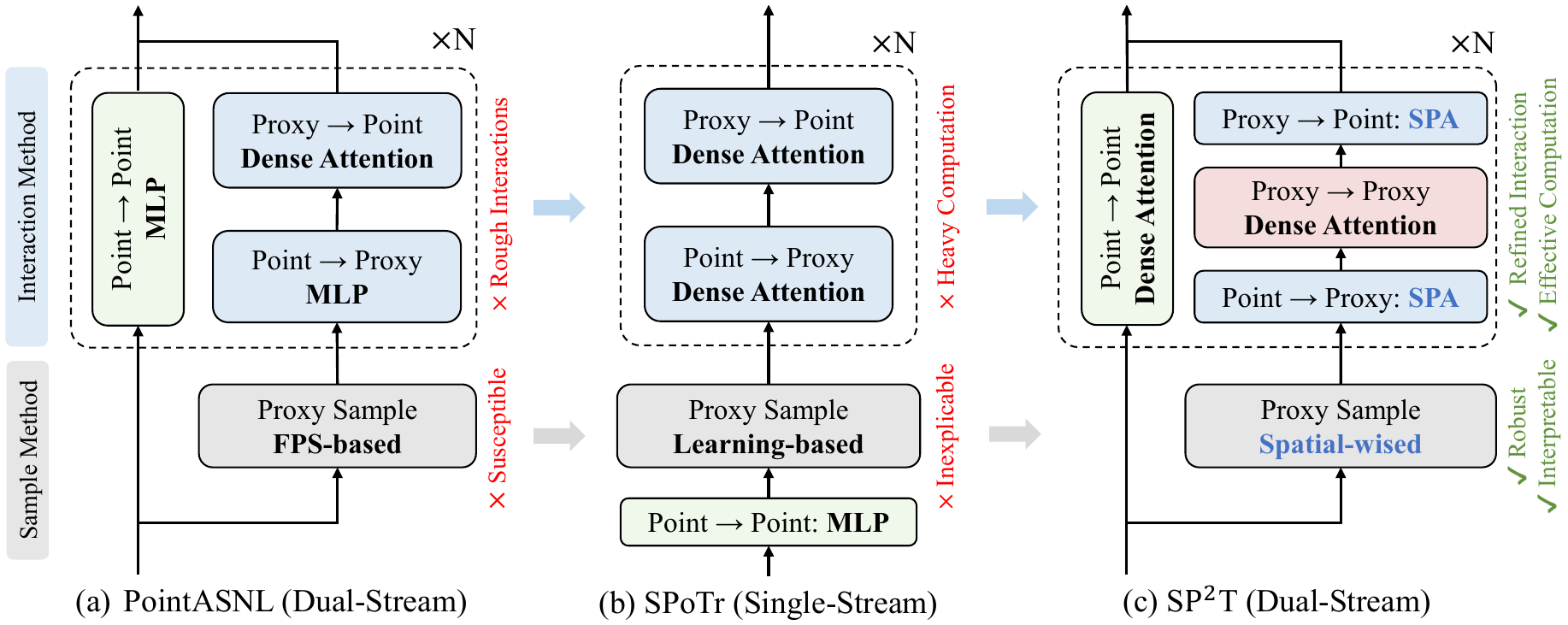}
    \vspace{-6mm}
    \caption{The comparison with other Proxy-based Methods. The improvement of SP$^2$T compared with other proxy-based method are threefold: (1) \textbf{Advances Sampling}, (2) \textbf{Efficiency Interaction} and (3) \textbf{Enhanced Dual-stream Framework}.}
    \vspace{-7mm}
    \label{fig:first}
\end{figure}

In recent years, proxies, which serve as key elements on the feature map or represent an abstraction of features for simplification, have attracted significant interest from the research community. Proxy-based methods are typically classified into global~\cite{wang2020linformer, jaegle2021perceiver, jaegle2021perceiverio, darcet2023vision, jaegle2021perceiver} and local approaches~\cite{lee2019set, huang2023vision, han2023agent, yan2020pointasnl, park2023self}, both with a global RF. However, global proxy methods have quadratic complexity, and the positional uncertainty of proxies adversely affects segmentation and detection tasks such as object boundary localization, making them unsuitable for point feature extraction.

In contrast, local proxy-based methods~\cite{lee2019set, huang2023vision, han2023agent, yan2020pointasnl, park2023self} employ a proxy sampling technique to establish the location of the proxy. They utilized the association to illustrate the link between data and proxy, with low-cost complexity and precise positioning ideal for extracting the feature of points. However, existing local proxy-based methods face challenges such as 1) inconsistent sampling from geometrically diverse points, 2) inefficient computation of proxy interactions, and 3) imbalanced integration of local and global information, as shown in Fig.~\ref{fig:first}. Initially, the sampling and association methods outlined are challenging to implement in geometrically diverse points, leading to an unstable spatial range of proxy. Secondly, handling sparse points introduces difficulties in optimizing the association process with cross-attention between proxies and points, particularly in terms of complexity and memory requirements with a considerable number of points. Third, existing proxy methods for points such as SPoTr~\cite{park2023self} and Fast PT~\cite{park2022fast} lose the balance between global and local information. In tasks like segmentation, global features are critical for class-level understanding, while local features define object boundaries; thus, striking a balance is crucial.

To address these challenges, we propose the Sparse Proxy Point Transformer (SP$^2$ T), a dual stream transformer with local proxies to understand the global receptive field. Inspired by MobileFormer~\cite{chen2022mobile}, PointASNL~\cite{yan2020pointasnl}, and others~\cite{mao2021dual, zhang2021dspoint}, SP$^2$T employs a dual stream structure: a proxy stream for compressed global representations and a point stream for detailed geometric features.

Secondly, for robust proxy sampling, we propose a spatial-wise binary selection method for an optimal proxy distribution. Compared to FPS, it reduces complexity, ensures uniformity, and mitigates information leakage from aggregation at the farthest point. We also introduce vertex-based association, aligning points to grid-vertex proxies equivalent to L-infinity KNN, enabling rapid performance.

Third, we propose Sparse Proxy Attention (SPA) for efficient cross-attention between sparsely connected proxies and points. SPA replaces matrix-based attention with linear complexity and employs Map-Reduce for parallel speedup. Additionally, Table-Based Relative Bias (TRB) is introduced to overcome traditional Relative Bias' limitations in sparse data, enhancing SPA's spatial perception.

In general, our improvement compared with other proxy-based method are following: (1) \textbf{Advances Sampling}: Replaces FPS and learning-based method with a spatial-wise method using vertex-based association, achieving explainability and scenario adaptability through various 3D scenes. (2) \textbf{Efficiency Interaction}: SPA reduces the computation of $O(nmd)$ of dense attention to $O(nkd+m^2d) (k=8,m<<n)$, facilitating the implementation of proxy-based methods on large 3D scenes. (3) \textbf{Enhanced Dual-stream Framework}: Optimized with attention-based local fusion to improve local extraction with global-local equilibrium.  (4) \textbf{SOTA Performance}: Extensive experiments indicate that SP$^{2}$T achieves state-of-the-art performance with acceptable latency (20~FPS in nuScenes~\cite{caesar2020nuscenes}) on indoor and outdoor 3D understanding benchmarks~\cite{dai2017scannet, rozenberszki2022language, armeni20163d, caesar2020nuscenes, behley2019semantickitti, sun2020scalability}, showing significant improvements (+3.8\% mIoU compared to SPoTr@S3DIS~\cite{armeni20163d}, +22.9\% mIoU compared to PointASNL@Sem.KITTI~\cite{behley2019semantickitti}) relative to other proxy-based point cloud methods.



\section{Related Works}

\nbf{3D Point Cloud Understanding.}
Currently, methods oriented towards 3D Point Cloud Understanding can be divided into three main categories: 2D projection-based~\cite{su2015multi,li2016vehicle,chen2017multi,lang2019pointpillars}, voxel-based~\cite{maturana2015voxnet,song2017semantic, zhou2018voxelnet, chen2023voxelnext, fan2022embracing, wang2023dsvt, zhang2024sparse, he2024scatterformer}, and point cloud-based methods~\cite{qi2017pointnet,qi2017pointnet++,zhao2019pointweb, qian2022pointnext, zhao2021point, wu2022point, wu2024point, lin2023meta} for down-stream task~\cite{zhang2025coddiff, lyu2023uedg, zhang2025awaretrack}. SP$^2$T is classified as a point-cloud-based network, which uses proxies to expand the receptive field of the point-cloud network significantly. Compared to PTv3~\cite{wu2024point}, the primary focus of PTv3 is serialization attention, which is a unique style of group attention~\cite{lin2023meta}. However, further expansion of the window size of serialization attention diverts the model’s focus, impairing its ability to capture precise features. In contrast, SPA uses proxies to maintain feature perception with a global receptive field. Additionally, our hybrid structure takes advantage of PTv3's serialization attention for capturing local patterns and SPA for facilitating global interactions.

\nbf{Proxy-based Methods.} 
Recently, proxy-based methods for feature extraction have gained considerable attention. These methods are categorized mainly into global proxy-based approaches~\cite{wang2020linformer, jaegle2021perceiver, jaegle2021perceiverio, darcet2023vision, park2022fast, jaegle2021perceiver} and local proxy-based approaches~\cite{lee2019set, huang2023vision, han2023agent, yan2020pointasnl, park2023self}. The proxy method faces challenges such as complexity, sampling, and association methods. Consider Register ViT~\cite{darcet2023vision} (global) and STViT~\cite{huang2023vision} (local) as case studies. The ViT register faces a complexity \(O(n^2)\), leading to inefficiencies with large data sets, while the STViT scratch window sampling, intended for 2D data, is not well equipped to handle the sparse nature of point clouds. Furthermore, PointASNL~\cite{yan2020pointasnl}, as a proxy-based point network, is severely limited to a crude point-proxy interaction design. In contrast, SP$^2$T employs a local proxy-based framework and examines the spatial-wise sampling and vertex-based association method for point clouds, using sparse proxy attention to accelerate the computation of proxy-point interactions. Sparse Proxy Attention efficiently establishes sparse associations between points and proxies, making it better suited for 3D environments. In addition, some proxy-based models for point clouds, such as Fast PT~\cite{park2022fast} and SPoTr~\cite{park2023self}, overlook the equilibrium between local and global information, resulting in poor performance in indoor datasets. SP$^2$T employs a network architecture that harmonizes global and local information, enabling proxy-based models to achieve competitive precision in outdoor and indoor data sets for the first time.

\section{Method}

\subsection{Overall Architecture}

\begin{figure*}[t]
    \centering
    \includegraphics[width=0.90\linewidth]{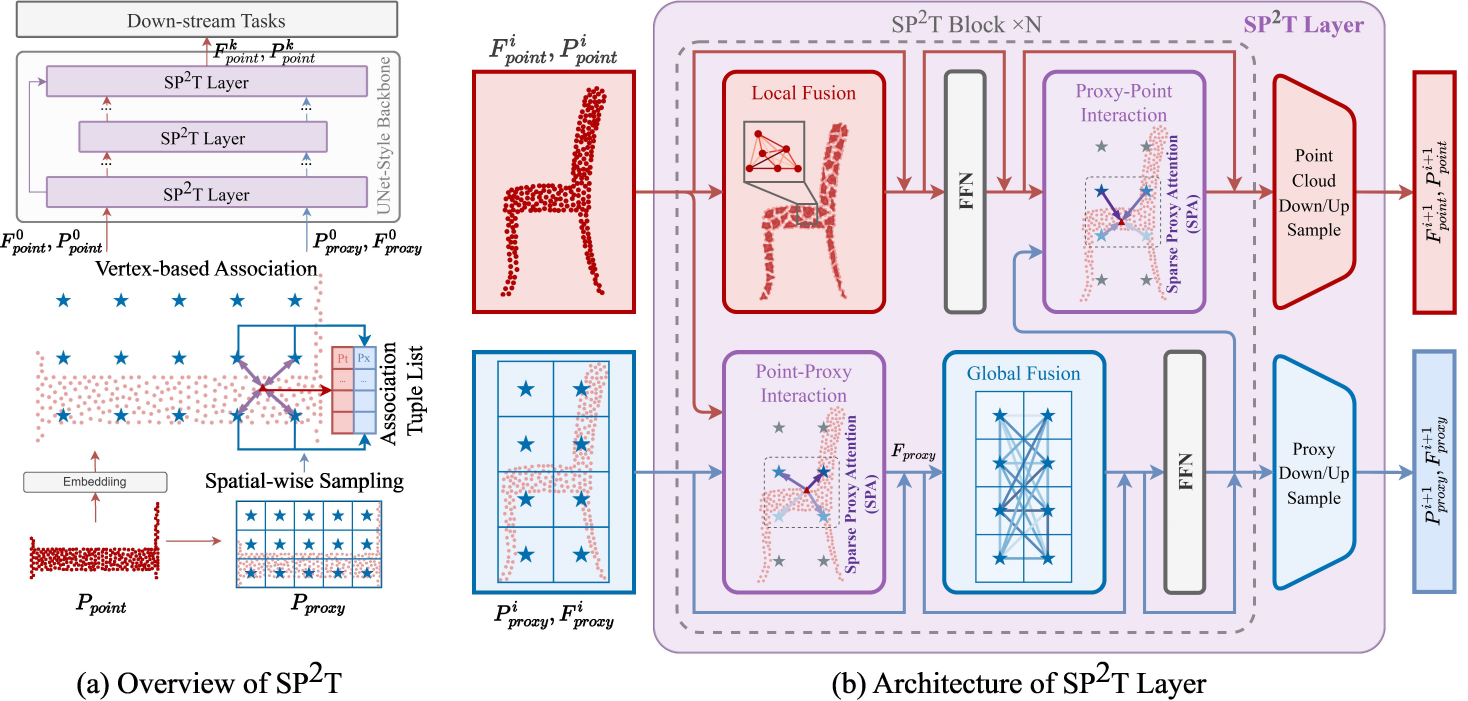}
    \vspace{-2mm}
    \caption{(a) Overview of SP$^2$T. Followed by PTv3, our model is a U-Net Style Transformer. (b) Architecture of SP$^2$T Layer. Every SP$^2$T layer comprises several SP$^2$T blocks, where both point and proxy must execute four essential operations: local fusion, point-proxy interaction, and global fusion combined with proxy-point interaction.}
    \vspace{-5mm}
    \label{fig:overview}
\end{figure*}

\nbf{SP$^2$T.} As illustrated in Fig.~\ref{fig:overview}, SP$^2$T is a dual-stream network specifically designed for points. Formally, the input points $P_{point} \in \mathbb{R}^{N \times 3}$ are processed through the point embedding module and the proxy initialization module to derive the feature of the point $F_{point}^{0} \in \mathbb{R}^{N \times C}$ and the proxy $F_{proxy}^{0} \in \mathbb{R}^{M \times C}$ and the position of the proxy $P_{proxy} \in \mathbb{R}^{M \times 3}$, where $N$ and $M$ is the number of points and proxy, and $C$ is the channel. The point embedding comprises several MLPs, while the proxy initialization component includes sampling, association, and several MLPs to synthesize $F_{proxy}^{0}$ from points associated with proxies. For the proxy feature $F_{Proxy}$, the model applies sinusoidal position embedding to define its feature as $F_{Proxy} = \text{Sinusoidal}(P_{proxy})$ after the proxy identifies its location $P_{proxy} \in \mathbb{R}^3$ using proxy sampling.

Afterward, the positions and features of both the points and proxies are fed into a multi-stage encoder-decoder network. Each SP$^2$TLayer is made up of several SP$^2$TBlock for the exaction of features and ends with an up/downsampling module for points $P_{point}^{i}, F_{point}^{i}$ and a proxy $P_{proxy}^{i}, F_{proxy}^{i}$ to facilitate the encoding and decoding processes, where $i$ is the idx of Each SP$^2$TLayer. Finally, the features and location of the points $P_{point}^{k}, F_{point}^{k}$ produced by the last layer $k$ of the decoder are utilized for downstream tasks. For network details, we report them in \emph{supplementary materials}.

\nbf{SP$^2$TBlock.} The SP$^2$TBlock is the foundational module in SP$^2$T, enabling distinct local feature aggregation for points and global feature aggregation for proxies. SP$^2$TBlock contains four distinct modules for feature extraction: local fusion, global fusion, point-proxy interaction, and proxy-point interaction module. Specifically, the local fusion module of the layer $i$ is designed to seamlessly integrate the local feature of the points $F_{proxy}^{i}$. In contrast, the global fusion module exceeds the limitations of the local receptive field for a comprehensive understanding of the global context. Subsequently, the point-proxy and proxy-point interaction modules combine the features of the points with the proxies $F_{proxy}^{i}$ using sparse proxy attention (SPA). These modules primarily enhance the exchange of information between the points and the proxy, enriching the features of the points with a detailed semantic context while maintaining the integrity of the local representation of the features of the proxies.

\subsection{Proxy Sampling and Association}

\nbf{Problem of FPS-based Proxy Sampling.} 
FPS-based proxy sampling utilizes the Furthest Point Sampling algorithm~\cite{qi2017pointnet++} to select a constant number of reference points from the points, as widely applied in PointNet~\cite{qi2017pointnet} and other methods. 
However, in our method, the FPS-based approach decreases precision. FPS-based sampling generates a scene-relevant proxy. The model might overfit such a proxy distribution during training, resulting in a segmentation output derived directly from the scene-relevant proxy rather than the point feature.

\begin{figure}[t]
    \centering
    \includegraphics[width=\linewidth]{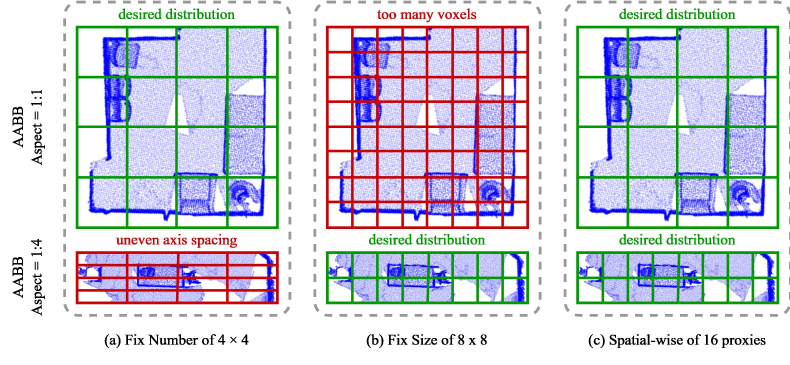}
    \vspace{-7mm}
    \caption{Comparison between different grip-based sampling methods and Spatial-wise method. (a) Fix Number of $4 \times 4$; (b) Fix Size of $8 \times 8$; (c) Spatial-wise sampling method by binary searching the square proxy spacing with a target proxy count \(16\).}
    \vspace{-6mm}
    \label{fig:grid-sampling-cmp}
\end{figure}

\nbf{Problem of Grid-based Proxy Sampling.}
Regarding grid-based proxy sampling methods, there are two primary approaches to sampling: one maintains a constant number of proxies, while the other regulates a specific proxy spacing. However, each of these methods has particular challenges. Firstly, the sampling method that keeps the number of proxies constant uses a predetermined quantity \(X_{voxel} \times Y_{voxel} \times Z_{voxel}\) within a uniform range. However, as illustrated in Fig.~\ref{fig:grid-sampling-cmp}~(a), since the aspect ratio of the generated proxies mirrors that of the axis-aligned bounding box (AABB) of the points, the proxies will appear stretched if the AABB has an extreme aspect ratio. This uneven distribution along different directions can adversely affect the uniformity of associations and feature fusion. Secondly, the spacing-fixed sampling method separates the aspect ratio of the proxy's spacing from the points by fixing the size of the proxy's spacing, enabling the uniformity of the proxies' distribution in space. However, as illustrated in Fig.~\ref{fig:grid-sampling-cmp}~(b), since the AABB's range of the points can fluctuate, the resultant number of proxies generated also varies, introducing significant uncertainty regarding the FLOPs of the model.

\nbf{Spatial-wise Proxy Sampling.}
We propose a spatial-wise proxy sampling method to address the issues mentioned above. Specifically, we introduced a binary search method for finding the proxy spacing. It is noted that, under equal proxy spacing along different axes, the number of generated proxies monotonically decreases as the proxy spacing increases. Therefore, given a desired range of proxy numbers, the binary search method can be used to find an appropriate proxy spacing that brings the number of generated proxies as close as possible to the required range. 

In practice, our sampling method efficiently discerns the ideal proxy spacing for a fixed number of proxies by considering the AABB sizes of various points. The technique is designed to maintain the proxy count within the bounds of $N_{max}$ and $N_{min}$, using a bisection approach to determine the optimal proxy spacing $L_p$. If the number of proxies exceeds $N_{max}$, the proxy spacing $L_p$ is reduced, and the total number of proxies is recalculated. In contrast, $L_p$ increases until the number of proxies reaches the desired range from $N_{max}$ to $N_{min}$. As shown in Fig.~\ref{fig:grid-sampling-cmp}~(c), this method autonomously selects the best proxy spacing based on the predetermined proxy count for different points depending on their AABB size. Compared to the other sampling methods in Fig.~\ref{fig:grid-sampling-cmp}~(a) and (b), the spatial-wise proxy sampling method guarantees an appropriate aspect ratio for proxy spacing and specifies a distinct number of proxies. The pseudo-code of spatial-wise sampling is shown in \emph{supplementary materials}.

\begin{figure}[t]
    \centering
    \includegraphics[width=\linewidth]{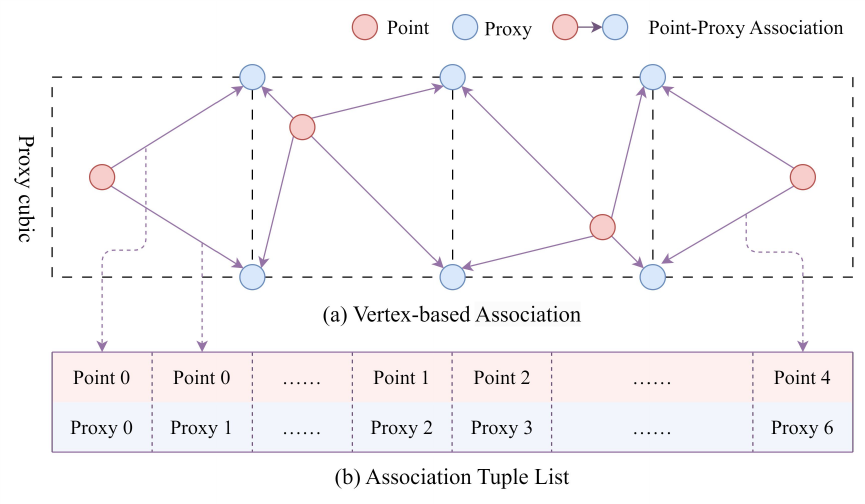}
    \vspace{-5mm}
    \caption{Overview of Vertex-based Association. In SPA, each point correlates with the proxies of the eight vertex on its respective proxy cubic.}
    \vspace{-5mm}
    \label{fig:asso}
\end{figure}

\nbf{Vertex-based Association.} 
A commonly adopted method for association uses the K-Nearest Neighbor (KNN)~\cite{taunk2019brief} method, which connects each data point with its $k$ nearest proxies. This approach provides considerable flexibility in defining the degree of correlation between the proxies and the points, thereby permitting modifications in the number of associations.
Despite this, the KNN method requires evaluating the distances between all points and proxies, which adds computational complexity. In KNN using the L-Infinity Norm with \(k = 8\), every point is linked to 8 proxies positioned at the corners of a cubic proxy within the AABB coordinate sphere. As the proxy is formed through grid sampling, its structural layout can be exploited to create an efficient association method, thereby reducing the need for intricate distance computations.
As illustrated in Fig.~\ref{fig:asso}, the initial transformation involves converting the coordinates of the points to their corresponding locations within the proxy grid. Then, the positions of the nearest eight proxies are determined by computing the floor and ceiling for each coordinate. Finally, the compilation of association tuples $as$ is achieved by merging all associations into a list.

\subsection{SPA for Point-Proxy Interaction}

The point-proxy interaction is the fundamental concept driving SP$^2$T. Unlike other works such as SAFDNet~\cite{zhang2024safdnet}, SP$^2$T facilitates the exchange of information between proxies and points through attention mechanisms and introduces sparse proxy attention to speed up the attention computation.

\nbf{Sparse Proxy Attention (SPA).}
Unlike traditional attention~\cite{vaswani2017attention}, the attention between the point and the proxy is sparse and discrete due to the vertex-based connections, leading to substantial computational demands with the typical attention operator. We implemented sparse proxy attention to address this issue and speed up information exchange between proxies and points. In practice, sparse proxy attention utilizes the points or proxy features as the query feature, while the other acts as the key and value features. Thus, it enables the aggregation of information from the point to the proxy and the feedback from the proxy back to the point.

\begin{figure}[t]
    \centering
    \includegraphics[width=\linewidth]{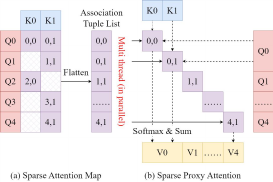}
    \vspace{-5mm}
    \caption{Overview of SPA calculation. SPA utilizes map-reduce computing based on association tuple lists to compute sparse attention between proxy and point. }
    \vspace{-5mm}
    \label{fig:attn}
\end{figure}

Take the SPA that transfers features from proxy to point as an example. As shown in Fig.~\ref{fig:attn}, to calculate the similarity between the proxy and the point, we denote the indices of the corresponding original points and proxy points in the \(i\)-th association tuple as \(as^{pt}_i, as^{px}_i\). The sparse attention module first computes the exponential similarity for each association group's query and each head's key features.

\vspace{-2mm}
\begin{equation}
    S^h_i = \exp\left( q^h_{as^{pt}_i} \cdot k^h_{as^{px}_i} / \sqrt{d} \right)
\end{equation}
\noindent where $S^h_i$ is the similarity of the $i$-th association of the head $h$, $q^h_{as^{pt}_i}$ and $k^h_{as^{px}_i}$ is the query $h$ of the head and key of \(as^{pt}_i, as^{px}_i\). $d$ is the dimension of the feature.

Building on this, we employ a map-reduce algorithm to obtain the sum of exponential similarity over each query. Finally, a division is performed to complete the sparse softmax and get the weights for each value.

\vspace{-2mm}
\begin{equation}
    W^h_{i} = \frac{S^h_i}{\sum_{j, as^{px}_j = as^{px}_i} S^h_j}
\end{equation}
\noindent where $S^h_i, S^h_j$ is the similarity of the $i$-th association and the $j$-th association of the same proxy. $W^h_{i}$ is the weight of the value after the sparse softmax. $as^{px}_{i}, as^{px}_{j}$ represents the index of the association proxy for points $i$ and $j$.

Finally, we adopt a similar map-reduce-based method to weigh the corresponding value features by the weights and association indices to obtain the output features.

\vspace{-2mm}
\begin{equation}
    o^{h}_i = \sum_{j, as^{px}_j = i} v^h_{as^{pt}_j}W^h_{i}
\end{equation}
\noindent where $v^h_{as^{pt}_j}$ is the association value $as^{pt}_j$ of the $h$-th head and $o^{h}_i$ is the attention output of the $h$-th head proxy $i$. The computational complexity of SPA is $O(knd)$, where $k$ is the association number and equal to 8, $n$ is the number of points, and $d$ is the channel dimension.

\begin{figure}[t]
    \centering
    \includegraphics[width=\linewidth]{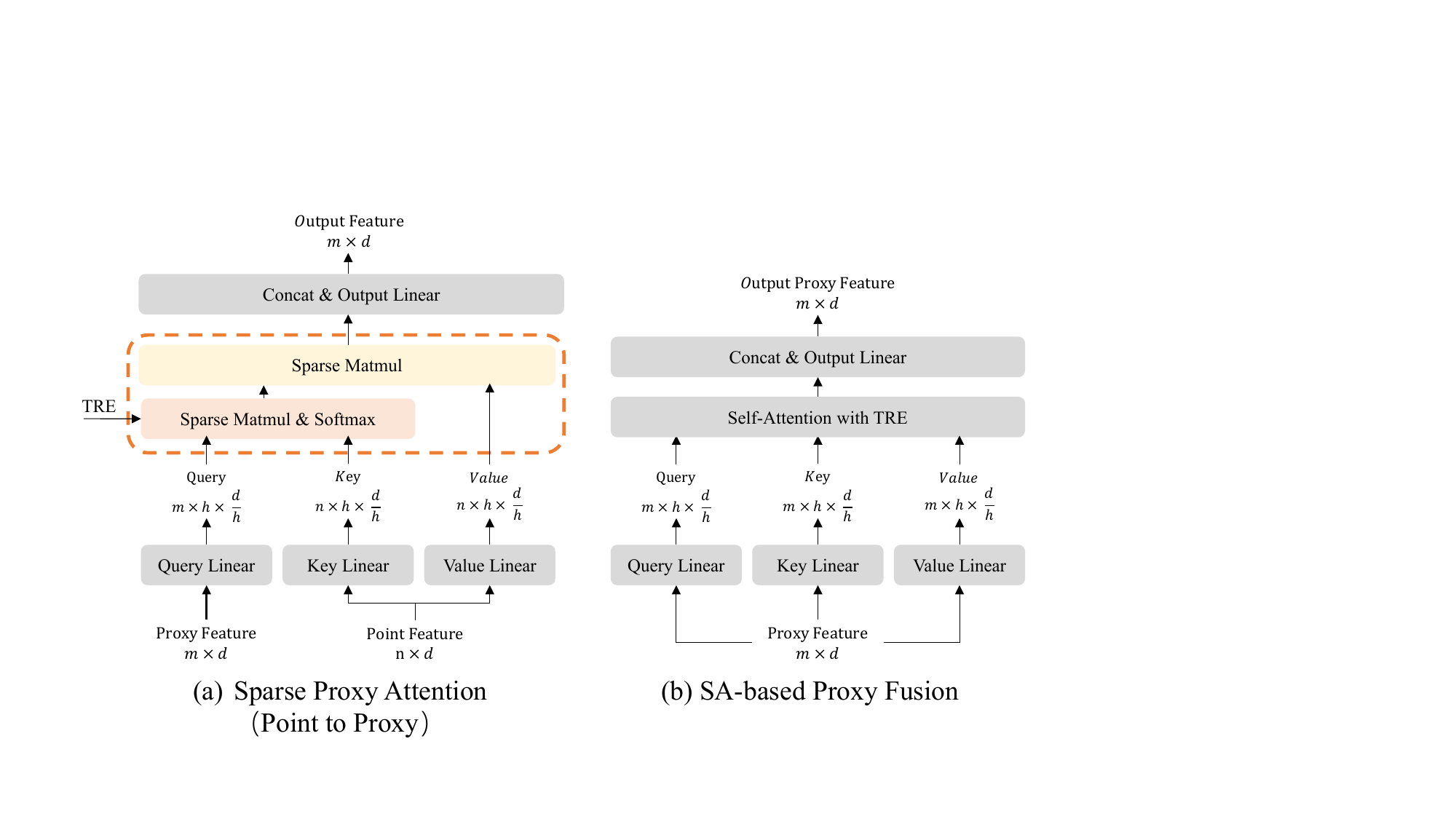}
    \vspace{-5mm}
    \caption{(a) The architecture of Sparse proxy Attention of point-proxy interaction. (b) The architecture of global fusion.}
    \vspace{-5mm}
    \label{fig:proxynet}
\end{figure}

\nbf{Table-based Relative Bias.}
\label{sec:RPE}
Inspired by the Swin Transformer~\cite{liu2021swin}, we found that integrating relative bias (RB) into the attention mechanism improves the network's proficiency in capturing the connections between the proxy and the points. However, the relative bias in the Swin Transformer is a learning parameter synchronized with the attention map. In contrast, in SPA, the relative bias must adapt to match the dimensions of the points and proxy.

We introduce Table-Based Relative Bias (TRB) to address the abovementioned issues. The TRB utilizes a lookup table \(T_{rpe}\) characterized by dimensions \(X_{rpe}, Y_{rpe}, Z_{rpe}\). The input relative positions are initially converted into normalized coordinates for the lookup table by applying a scaling factor \(s_{rpe}\). The biases are then derived at these coordinates via trilinear interpolation. Relative distances beyond the permissible range are clamped to valid numerical values. In terms of form, the formula for TRB is as follows.
\vspace{-2mm}
\begin{equation}
    \operatorname{TRB}(x) = \operatorname{TGS}\left(T_{rpe}, \operatorname{clamp}(s_{rpe}x, -1, 1) \right)
\end{equation}

\noindent where $\operatorname{TGS}$ is the function of the TrilinearGridSample and $x$ is the distance between the proxy and the point. $\mathop{clamp}$ is the function of the clamp number.

Finally, the similarity formula with TRB is shown below.
\vspace{-2mm}
\begin{equation}
    S^h_i = \exp\left( q^h_{as^{pt}_i} \cdot k^h_{as^{px}_i} / \sqrt{d} \right) + 
            \operatorname{TRB}^h\left( p^{pt}_{as^{pt}_i} - p^{px}_{as^{px}_i} \right)
\end{equation}

\noindent where $S^h_i$ is the similarity of the $i$-th association of the head $h$, $q^h_{as^{pt}_i}$ and $k^h_{as^{px}_i}$ is the query of the head $h$ and key of \(as^{pt}_i, as^{px}_i\). $d$ is the dimension of the feature. $p^{pt}_{as^{pt}_i}, p^{px}_{as^{px}_i}$ is the position of the point and proxy.

Furthermore, the feature dimension of the lookup table $T_{rpe}$ is configured to match the number of heads in the sparse attention mechanism. This setup allows parallel execution of interpolation sampling and integration into similarity for each head. Furthermore, the initialization of $T_{rpe}$ used a Gaussian distribution with scaled variance, which initially encourages the network to focus on the proximate features.

\nbf{Structure of SPA.}  Fig.~\ref{fig:proxynet}~(a) illustrates the SPA structure with Point-to-Proxy. Similarly to MHSA~\cite{vaswani2017attention}, SPA incorporates four additional linear layers designed to scale down and increase the features.

\begin{table}[t]
    \begin{minipage}{\linewidth}
        \centering
            \resizebox{1.0\linewidth}{!} {
            \tablestyle{2.8pt}{1.1}
            \begin{tabular}{lcccccc}\toprule
Outdoor Sem. Seg. &\multicolumn{2}{c}{nuScenes~\cite{caesar2020nuscenes}} &\multicolumn{2}{c}{Sem.KITTI~\cite{behley2019semantickitti}} &\multicolumn{2}{c}{Waymo Val~\cite{sun2020scalability}} \\\cmidrule(lr){2-3} \cmidrule(lr){4-5} \cmidrule(lr){6-7} 
Methods &Val &Test &Val &Test &mIOU &mAcc  \\\midrule
 PointASNL~\cite{yan2020pointasnl} & - & - & 48.8 & - & - & - \\
 MinkUNet~\cite{choy20194d} &73.3 & - & 63.8 & - &65.9 &76.6 \\
 SphereFormer~\cite{lai2023spherical} & 78.4 & 81.9 & 67.8 & 74.8 & 69.9 & - \\
 PTv2~\cite{wu2022point} &80.2 & 82.6 & 70.3 & 72.6 &70.6  &80.2 \\
 PTv3~\cite{wu2024point} & 80.4 & 82.7 & 70.8 & 74.2  & 71.3 &80.5 \\
 \rowcolor{gray!20}
 SP$^2$T & \textbf{81.2} & \textbf{83.4} &  \textbf{71.7}  &  \textbf{75.4} &\textbf{71.9} &\textbf{82.5} \\
\bottomrule
\end{tabular}
            }
            \vspace{-2.5mm}
            \caption{Outdoor semantic segmentation.}\label{tab:outdoor_sem_seg}
            \vspace{2mm}
    \end{minipage} \\
    \begin{minipage}{\linewidth}
        \centering
            \resizebox{1.0\linewidth}{!} {
            \tablestyle{2.8pt}{1.1}
            \begin{tabular}{lcccccccc}\toprule
Waymo~\cite{sun2020scalability} Det. &  &\multicolumn{2}{c}{Vehicle L2} &\multicolumn{2}{c}{Pedestrian L2} &\multicolumn{2}{c}{Cyclist L2} & Mean L2\\\cmidrule(lr){3-4} \cmidrule(lr){5-6} \cmidrule(lr){7-8} \cmidrule(lr){9-9} 
Methods & \# &mAP &mAPH &mAP &mAPH &mAP &mAPH & mAPH  \\\midrule
CenterPoint~\cite{yin2021center} & 1 & 66.7 & 66.2 & 68.3 & 62.6 & 68.7 & 67.6 & 65.5 \\
PillarNet~\cite{lang2019pointpillars} & 1 & 70.4 & 66.9 & 71.6 & 64.9 & 67.8 & 66.7 & 67.2\\
PTv3~\cite{wu2024point}    & 1 & 71.2 & 70.8 & 76.3 & 70.4 & 71.5 & 70.4 & 70.5 \\
\rowcolor{gray!20}
SP$^2$T & 1 & \textbf{72.1} & \textbf{71.6} & \textbf{76.7} & \textbf{71.5} & \textbf{74.2} & \textbf{73.3} & \textbf{72.1}  \\\midrule
CenterPoint~\cite{yin2021center} & 2 & 67.7 & 67.2 & 71.0 & 67.5 & 71.5 & 70.5 & 68.4\\
PillarNet~\cite{lang2019pointpillars} & 2 & 71.6 & 71.6 & 74.5 & 71.4 & 68.3 & 67.5 & 70.2\\
PTv3~\cite{wu2024point}    & 2 & 72.5 & 72.1 & 77.6 & 74.5 & 71.0 & 70.1 & 72.2 \\
\rowcolor{gray!20}
SP$^2$T & 2 & \textbf{72.9} & \textbf{72.6} & \textbf{78.1} & \textbf{75.2} & \textbf{75.1} & \textbf{74.2} & \textbf{74.0}  \\
\bottomrule
\end{tabular}
            }
            \vspace{-2.5mm}
            \caption{Waymo object detection.}\label{tab:waymo_det}
            \vspace{-7mm}
    \end{minipage} \\
\end{table}

\subsection{Local and Global Fusion}

\nbf{Local Fusion.}
The local fusion module, implemented using cross-architecture implementations (MinkUnet/Ptv3), performs local feature aggregation via point interactions. PTv3 was selected for its advantages in local fusion. In particular, SP$^2$T employs point-cloud serialization attention to facilitate the local fusion of points. 

\nbf{Global Fusion.}
Global fusion extends proxy receptive fields through TRB-driven self-attention, with experimental validation that confirms its critical role in SP$^2$T. Fig.~\ref{fig:proxynet}~(b) illustrates the global fusion using a self-attention with TRB, which facilitates the exchange of information across all proxies, allowing each to have direct access to the global context.

\section{Experiments}

\subsection{Result of Downstream Tasks}

This section will compare our model with other SOTA models in various downstream tasks. For detailed model and training settings, we report them in \emph{supplementary material}. 

\nbf{Outdoor Semantic Segmentation.}
The validation and test results of our model for the nuScenes~\cite{caesar2020nuscenes}, Semantic KITTI~\cite{behley2019semantickitti} and Waymo~\cite{sun2020scalability} benchmarks with other methods~\cite{choy20194d, Cheng2021af2s3net, yan20222dpass, lai2023spherical, wu2022point, wu2024point, yan2020pointasnl} are enumerated in Tab.~\ref{tab:outdoor_sem_seg}. Our model outperformed existing models, obtaining the highest results for both datasets. Specifically, relative to SphereFormer, we achieved a 2.8\%, 3.9\%, and 2.0\% boost on scenes, SemanticKITTI, and Waymo. In contrast to the proxy-based method, such as PointASNL~\cite{yan2020pointasnl}, SP$^2$T exhibits a significant improvement in SemanticKITTI by 22.9\%, underscoring the critical role of sparse proxy attention.

\begin{table}[t]
    \begin{minipage}{\linewidth}
        \centering
            \resizebox{\linewidth}{!} {
            \tablestyle{1.3pt}{1.1}
            \begin{tabular}{lcccccc}
\toprule
Indoor Sem. Seg.&\multicolumn{2}{c}{ScanNet~\cite{dai2017scannet}} &\multicolumn{2}{c}{ScanNet200~\cite{rozenberszki2022language}} &\multicolumn{2}{c}{S3DIS~\cite{armeni20163d}} \\\cmidrule(lr){2-3} \cmidrule(lr){4-5} \cmidrule(lr){6-7}
Methods &Val &Test$^\dag$ &Val &Test$^\dag$ &Area5 &6-fold \\\midrule
  \multicolumn{7}{l}{Conv-based Method:} \\
 MinkUNet~\cite{choy20194d} &72.2 &73.6 &25.0 &25.3 &65.4 &65.4 \\\midrule
 \multicolumn{7}{l}{Transformer-based Method:} \\
  PTv2~\cite{wu2022point} &75.4 & - &30.2 &- &71.6 &73.5 \\
  OctFormer~\cite{wang2023octformer} &75.7 & 70.7 &31.9 &32.6 &- &- \\
  Swin3D~\cite{yang2023swin3d} &76.4 & 71.4 & - &- & 72.5 &76.9 \\
 PTv3~\cite{wu2024point} & 77.5 &73.6 & 35.2 & 34.0 & 73.4 & 77.7 \\\midrule
 \multicolumn{7}{l}{Proxy-based Method:} \\
  PointASNL~\cite{yan2020pointasnl} & 63.0 & - & - & - & 68.7 & - \\
 Fast PT~\cite{park2022fast} & 72.1 & - & - & - & 70.1 & - \\
 SPoTr~\cite{park2023self} & - & - & - & - & 70.8 & - \\
 \rowcolor{gray!20}
 SP$^2$T &\textbf{78.7} & \textbf{74.9} & \textbf{37.0} & \textbf{35.2} & \textbf{74.6} & \textbf{79.7}  \\
\bottomrule
\end{tabular}

            }
            \vspace{-2.5mm}
            \caption{Indoor semantic segmentation. $^\dag$The result in test set of ScanNet and ScanNet200 use the PTv3's \emph{TTA of Val Set}. }\label{tab:indoor_sem_seg}
            \vspace{2mm}
    \end{minipage} \\
    \begin{minipage}{\linewidth}
            \centering
                \resizebox{\linewidth}{!} {
                \tablestyle{2.8pt}{1.1}
                \begin{tabular}{lccccccc}\toprule
Id. Ins. Seg.&\multicolumn{3}{c}{ScanNet Val~\cite{dai2017scannet}} &\multicolumn{3}{c}{ScanNet200 Val~\cite{rozenberszki2022language}} \\\cmidrule(lr){2-4} \cmidrule(lr){5-7}
Baseline~\cite{jiang2020pointgroup} &mAP$_{25}$ &mAP$_{50}$ &mAP &mAP$_{25}$ &mAP$_{50}$ &mAP \\\midrule
 MinkUNet~\cite{choy20194d} &72.8 &56.9 &36.0 &32.2 &24.5 &15.8 \\
 PTv2~\cite{wu2022point} &76.3 &60.0 &38.3 &39.6 &31.9 &21.4 \\
 PTv3~\cite{wu2024point} &77.5 & 61.7 & 40.9 & 40.1 & 33.2 & 23.1 \\
 \rowcolor{gray!20}
 SP$^2$T & \textbf{78.8} & \textbf{62.9} & \textbf{42.2} & \textbf{41.6} & \textbf{34.3} & \textbf{24.7} \\
\bottomrule
\end{tabular}
                }
                \vspace{-2.5mm}
                \caption{Indoor instance segmentation.}\label{tab:indoor_ins_seg}
                \vspace{-8mm}
    \end{minipage} \\
\end{table}

\nbf{Outdoor Object Detection.} 
In Tab.~\ref{tab:waymo_det}, we present a comparison of SP$^2$T with other single-stage 3D detectors~\cite{yin2021center, lang2019pointpillars, wu2024point} on the Waymo Object Detection benchmark~\cite{sun2020scalability}. Our baseline is CenterPoint~\cite{yin2021center} followed by PTv3~\cite{wu2024point}. Our model consistently surpasses sparse convolutional~\cite{yin2021center, lang2019pointpillars} and transformer-based detectors~\cite{wu2024point}, demonstrating enhancements over recent SOTA. Notably, SP$^2$T surpasses PTv3~\cite{wu2024point} by 1.6\% using a single input frame and retains a 1.8\% lead in multi-frame scenarios.

\nbf{Indoor Semantic Segmentation.}
In Tab.~\ref{tab:indoor_sem_seg}, we compare our validation and test outcomes with other approaches~\cite{wu2022point, wu2024point, zhao2021point, wang2023octformer, qian2022pointnext, choy20194d, yang2023swin3d} on the ScanNet v2~\cite{dai2017scannet} and ScanNet200~\cite{rozenberszki2022language} benchmarks. This includes Area 5 and the 6-fold cross-validation~\cite{qi2017pointnet} evaluated on S3DIS~\cite{armeni20163d}. SP$^2$T achieved a performance boost of 1.2\% in the ScanNet and showed enhancements of 1.8\% in the ScanNet200, compared to PTv3~\cite{wu2024point}. In S3DIS, SP$^2$T achieved performance gains of 1.2\% and 2.0\% for Area 5 and 6-folds over PTv3. Compared to proxy-based approaches like PointASNL~\cite{yan2020pointasnl}, Fast PT~\cite{park2022fast} and  SPoTr~\cite{park2023self},  demonstrates a 15.7\% over PointASNL and 5.6\% over Fast PT on ScanNet, and shows gains of 5.9\% over PointASNL, 4.5\% over Fast PT and 3.8\% over SPoTr on S3DIS, proofing the breakthrough of SP$^2$T in proxy-based model design.
In the ScanNet and ScanNet 200 test data sets, the results of the models are presented using the augmentation of test time within the val set and do not use additional methods such as over-segmentation~\cite{nekrasov2021mix3d, schult2023mask3d, yang2023swin3d}, aggregating results from several trained models~\cite{lai2022stratified}, or employing more data during training. More details are provided in \emph{supplementary material}.

\begin{table}[t]
    \centering
        \resizebox{\linewidth}{!} {
        \tablestyle{3pt}{1.1}

\begin{tabular}{cl|cccc}
\toprule
No. & Methods & mIOU & mAcc & allAcc & Times/ms\\ \midrule
1 & PTv3 (Baseline) & 77.51 & 85.16 & 92.24 & 61 \\
2 & PTv3 with 1.1x channels & 77.64 ($\uparrow$ 0.13) & 85.41 & 92.28 & 73 \\\midrule
3 & \#1 + Pool-based PPI& 77.47 ($\downarrow$ 0.04) & 85.09 & 92.15 & 64 \\
4 & + Global Fusion  & 77.61 ($\uparrow$ 0.10) & 85.08 & 92.07 & 69 \\\midrule[0.1pt]
5 & \#1 + Conv-based PPI&  77.70 ($\uparrow$ 0.19) & 85.87 & 92.20 & 65 \\
6 & + Global Fusion  &  78.17 ($\uparrow$ 0.66) & 86.13 & 92.16 & 70 \\\midrule[0.1pt]
7 & \#1 + Attn-based PPI w/o TRB&  77.83 ($\uparrow$ 0.32) & 85.93 & 92.14 & 68 \\
8 & + Global Fusion  &  78.24  ($\uparrow$ 0.73)& 86.22 & 92.27 & 73 \\\midrule[0.1pt]
9 & \#1 + Attn-based PPI w/ TRB &  78.32 ($\uparrow$ 0.81) & 86.16 & 92.37 & 69 \\
\rowcolor{gray!20}
10 & + Global Fusion  & \textbf{78.71 ($\uparrow$ 1.20)} & \textbf{86.23} & \textbf{92.51} & 74 \\
\bottomrule
\end{tabular}
        }
        \vspace{-3mm}
        \caption{Ablation study about structure of SP$^2$T. PPI means Point-Proxy Interaction.}\label{tab:ablation_structure}
        \vspace{-5mm}
\end{table}

\nbf{Indoor instance Segmentation.}
In Tab.~\ref{tab:indoor_ins_seg}, we provide validation outcomes for our model on the instance segmentation benchmarks of ScanNet v2~\cite{dai2017scannet} and ScanNet200~\cite{rozenberszki2022language}. Performance metrics displayed include mAP, mAP$_{25}$, and mAP$_{50}$, and are compared to various state-of-the-art (SOTA) backbones. Following PTv3~\cite{wu2024point}, we use the instance segmentation framework utilizing PointGroup~\cite{jiang2020pointgroup} in all evaluations, altering only the backbone. Compared to PTv2~\cite{wu2022point}, we achieved enhancements of 2.5\% in mAP$_{25}$, 2.9\% in mAP$_{50}$, and 3.9\% in mAP on ScanNet. Furthermore, in ScanNet200, there were gains of 2.0\% in mAP$_{25}$, 2.4\% in mAP$_{50}$, and 3.3\% in mAP.

\subsection{Ablation Study}
\nbf{The structure of SP$^2$T.}
In Tab.~\ref{tab:ablation_structure}, the structure ablation experiments of SP$^2$T are presented. The first two rows of Tab.~\ref{tab:ablation_structure} detail the baseline PTv3~\cite{wu2024point}, along with a version of PTv3 that matches the final size of the SP$^2$T model. This analysis shows that simply enlarging PTv3's parameters does not significantly enhance metrics because of limited receptive fields. Therefore, SP$^2$T's improved performance is not due to having more parameters. The 3 and 4 rows of Tab.~\ref{tab:ablation_structure} focus on a proxy model employing pool-based proxy-point interactions (PPI), revealing that poor-based PPI between agent and point cloud can degrade performance. Furthermore, rows 5 and 6 illustrate a model with convolutional PPI via TRB, compared to one using only TRB, showing that convolution improves feature extraction in the model, unlike pooling. Finally, rows 7 to 10 of Tab.~\ref{tab:ablation_structure} examine models with and without positional coding in attentional PPI. These findings indicate that although using attention or TRB alone is beneficial, their integration notably boosts performance, aligning with previous attention studies~\cite{han2022survey, khan2022transformers}. Moreover, ablation tests involving global fusion show enhanced performance across interaction modes, thanks to the global fusion capability to broadly extend the agent model's receptive field.

\nbf{The number of proxy and association.}
Tab.~\ref{tab:ablation_proxy_number} rows~1 to 6 display how varying the number of proxies affects SP$^2$T. The results indicate that the model demonstrates limited sensitivity to changes in the number of proxies, with the accuracy showing only minor fluctuations once the threshold (120) is reached. In contrast, rows~7 to 10 of Tab.~\ref{tab:ablation_proxy_number} address the study of association counts in SP$^2$T. The results reveal that the model reacts more notably to the different numbers of associations. Neither excessively high nor low association counts have a more pronounced impact on segmentation accuracy than variations in proxy numbers.

\begin{table}[t]
    \centering
        \tablestyle{2.8pt}{1.1}

\begin{tabular}{c|cccc|cccc}
\toprule
No. & Proxy & Asso. & Sample & E.P. & mIou & mAcc & allAcc & Time/ms  \\\midrule
1 & 40 & 8  & Wise. & \usym{2714}& 77.74 & 86.39 & 92.05 & \textbf{65} \\
2 &80 & 8 & Wise. & \usym{2714}& 77.99 & 86.11 & 92.17 & 68 \\
3 &120 & 8 & Wise.& \usym{2714}& 78.41 & 86.22 & 92.43 & 71 \\
\rowcolor{gray!20}
4 &160 & 8 & Wise.& \usym{2714}& \textbf{78.71} & 86.23 & \textbf{92.51} & 74 \\
5 &320 & 8 & Wise.& \usym{2714}& 78.68 & 86.15 & 92.31 & 92 \\
6 &512 & 8 & Wise.& \usym{2714}& 78.52 & \textbf{86.60} & 92.30 & 120 \\\midrule
7 &160 & 4  & Wise.& \usym{2714}& 77.99 & 86.61 & 92.22 & 69\\
\rowcolor{gray!20}
8 &160 & 8  & Wise. & \usym{2714}& \textbf{78.71} & 86.23 & \textbf{92.51} & 74 \\
9 &160 & 12 & Wise.& \usym{2714}& 77.90 & 85.77 & 92.19 & 97 \\
10 &160 & 16 & Wise.& \usym{2714} & 78.10 & \textbf{86.99} & 92.40  & 158 \\\midrule
11 &160 & 8  & FPS & \usym{2714}& 78.04 & 86.13 & 92.20 & 83 \\
12 &160 & 8 & Fix-Num. & \usym{2714}& 78.15 & 86.20 & 93.27 & 74 \\
13 &160 & 8 & Fix-Size & \usym{2714}& OOM & OOM & OOM & /  \\
\rowcolor{gray!20}
14 &160 & 8 & Wise. & \usym{2714}& \textbf{78.71} & \textbf{86.23} & \textbf{92.51}  & 74 \\\midrule
15 &160 & 8 & Wise.& \usym{2717} & 78.03   & 86.17 & 92.36  & 81 \\
\rowcolor{gray!20}
16 &160 & 8  & Wise. & \usym{2714} & \textbf{78.71} & \textbf{86.23} & \textbf{92.51} & 74 \\
\bottomrule
\end{tabular}

        \vspace{-2mm}
        \caption{Ablation study about hyper-parameter of SP$^2$T in ScanNet. The baseline structure of these models is Tab.~\ref{tab:ablation_structure}~\#10. Asso. means the number of associations. Wise. means Spatial-wise sampling. E.P. means Empty Proxies. OOM means out of memory. }\label{tab:ablation_proxy_number}
        \vspace{-2mm}
\end{table}

\begin{table}[t]
        \begin{minipage}{\linewidth}
            \centering
                \tablestyle{4.5pt}{1.1}
                \begin{tabular}{lcccc}
\toprule
Indoor Sem. Seg.&\multicolumn{2}{c}{ScanNet~\cite{dai2017scannet}} &\multicolumn{2}{c}{ScanNet200~\cite{rozenberszki2022language}}  \\\cmidrule(lr){1-1} \cmidrule(lr){2-3} \cmidrule(lr){4-5} 
Methods &Val &Test &Val &Test \\\midrule
 MinkUnet~\cite{choy20194d} &72.2 &73.4 &25.0 &25.3 \\
 SP$^2$T w/ MinkUnet & 75.7($\uparrow$3.5) & - & 30.3($\uparrow$5.3) &-  \\\midrule
 PTv3~\cite{wu2024point} & 77.5 & 73.6 & 35.2 & 34.0  \\
 SP$^2$T w/ PTv3 &78.7($\uparrow$1.2) & 74.9($\uparrow$1.3) & 37.0($\uparrow$1.8) & 35.2($\uparrow$1.2)  \\
\bottomrule
\end{tabular}
                \vspace{-2.5mm}
                \caption{Transfer study of local fusion.}\label{tab:appendix_transfur_study}
                \vspace{-5mm}
    \end{minipage} \\
\end{table}

\nbf{Proxy Sampling Method.}
Tab.~\ref{tab:ablation_proxy_number} rows~11 to 14 provides the results of the ablation study on proxy sampling methods. The experiment indicates that the spatial-wise sampling method yields the best model accuracy among various sampling methods. The Fix Number and Fix Size-based methods struggle with geometrically diverse point regarding the FPS-based methods, reducing accuracy and failure training. In contrast, the spatial-wise sampling approach can flexibly determine the proxy spacing and improve performance. The more discussion and visualization is provided in \emph{supplementary materials} regarding the FPS-based method.

\nbf{Empty Proxies.}
Unlike image, grid sampling in a point cloud unavoidably results in empty proxies, which lack any point cloud association. Tab.~\ref{tab:ablation_proxy_number} rows~15 to 16 compares SP$^2$T accuracy with and without considering empty proxies during global fusion. In the experiment, empty proxies are initialized with zero features and participate in global fusion to encode spatial voids. The experiment indicates that empty proxies enhance the accuracy of the model. The experiment proves that `nothing' is also critical information in the point cloud. Although empty proxies lack points, the absence itself conveys spatial structure information, thereby contributing to an understanding of overall spatial features for points.

\nbf{Transfer Study of Local Fusion.}
We also use MinkUnet~\cite{choy20194d} as the local fusion module to test the portability of our design. Tab.~\ref{tab:appendix_transfur_study} shows the performance of MinkUnet~\cite{choy20194d} and PTv3~\cite{wu2024point} and its variants with our proposed sparse proxy attention mechanism on the Scannet~\cite{dai2017scannet} and Scannet200~\cite{rozenberszki2022language} datasets. The experimental results demonstrate that our network structure is plug-and-play and can integrate seamlessly with various local fusion networks, enhancing the receptive field and boosting the network's performance.

\subsection{Visualization}

In this section, we illustrate the attention map and latency of SPA. More visualization and failed cases are provided in \emph{supplementary materials}.

\begin{figure}[t]
    \centering
    \includegraphics[width=\linewidth]{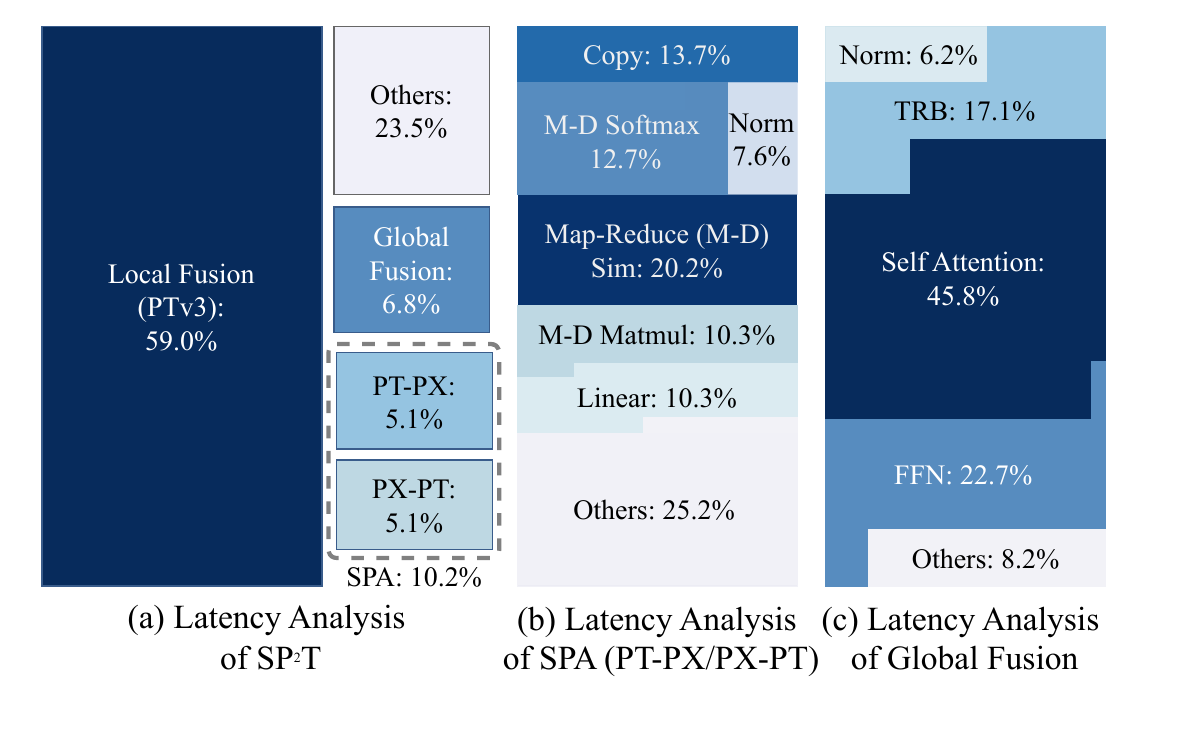}
    \vspace{-6mm}
    \caption{Visualization of the latency of SP$^2$T, Sparse Proxy Attention, and Global Fusion. The latency of local and global fusion includes the latency of FFN.}
    \vspace{-5mm}
    \label{fig:gpu_latency}
\end{figure}

\nbf{Visualization of Latency.}
Fig.~\ref{fig:gpu_latency}~(a) presents the latency tree map for our model. It reveals that SPA and global fusion contribute 17\% to overall latency. The latency of local and global fusion includes the latency of FFN. Additionally, Fig.~\ref{fig:gpu_latency}~(b)~\&~(c) display the latency tree map of each operator in SPA and global fusion. The findings indicate that SPA with global fusion significantly improves performance without imposing substantial delay.

\begin{figure}[t]
    \centering
    \includegraphics[width=\linewidth]{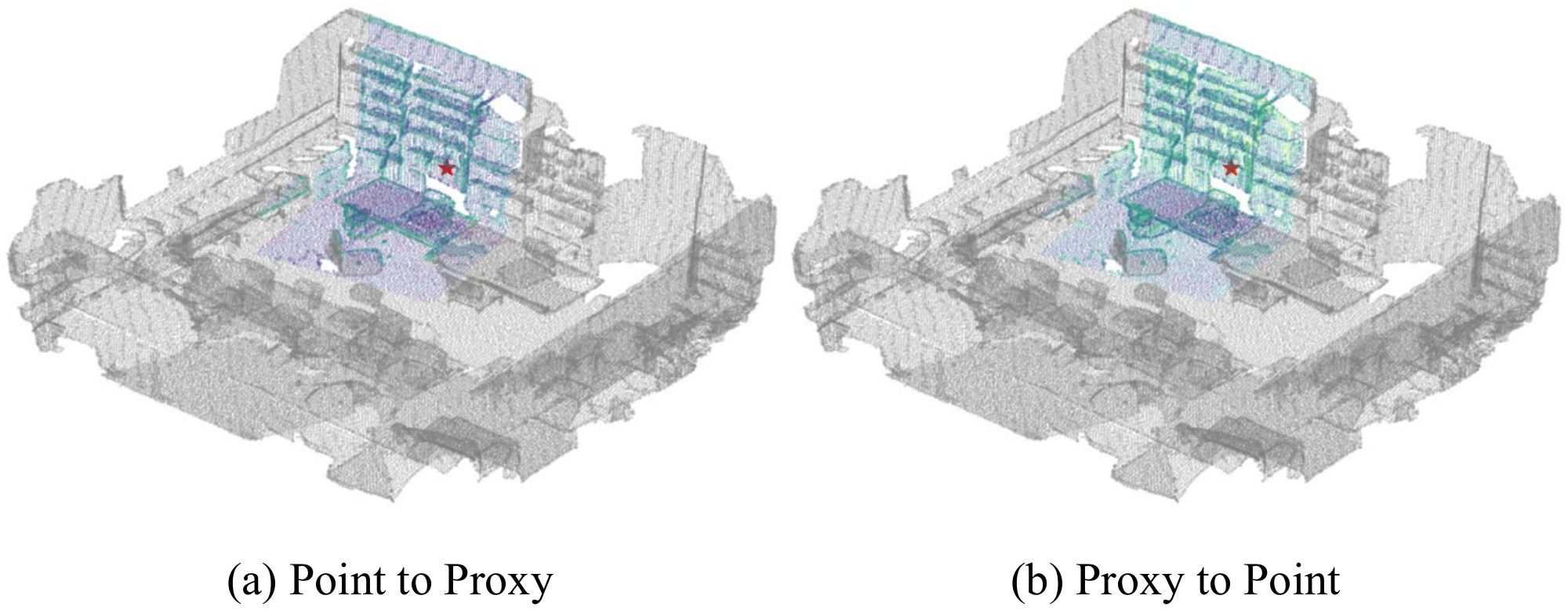}
    \vspace{-5mm}
    \caption{Visualization of Point-Proxy and Proxy-Point Attention Map of SPA. The red star represents the current proxy.}
    \vspace{-1mm}
    \label{fig:sparse-attn-vis}
\end{figure}

\nbf{Visualization of SPA.}
Fig.~\ref{fig:sparse-attn-vis} depicts the attention map in SPA, focusing on the interaction between the point and the proxy. Specifically, Fig.~\ref{fig:sparse-attn-vis}~(a) highlights the attention map from point to proxy, and Fig.~\ref{fig:sparse-attn-vis}~(b) showcases the return path from proxy to point. Examining the visualization reveals that the SPA's semantic mining capability is evident. In this visualization, as the chosen proxy approaches the bookshelf, the point cloud and the proxy's focus show a stronger affinity for the bookshelf than the table, showcasing the SPA's semantic representation ability.

\begin{figure}[t]
    \centering
    \includegraphics[width=\linewidth]{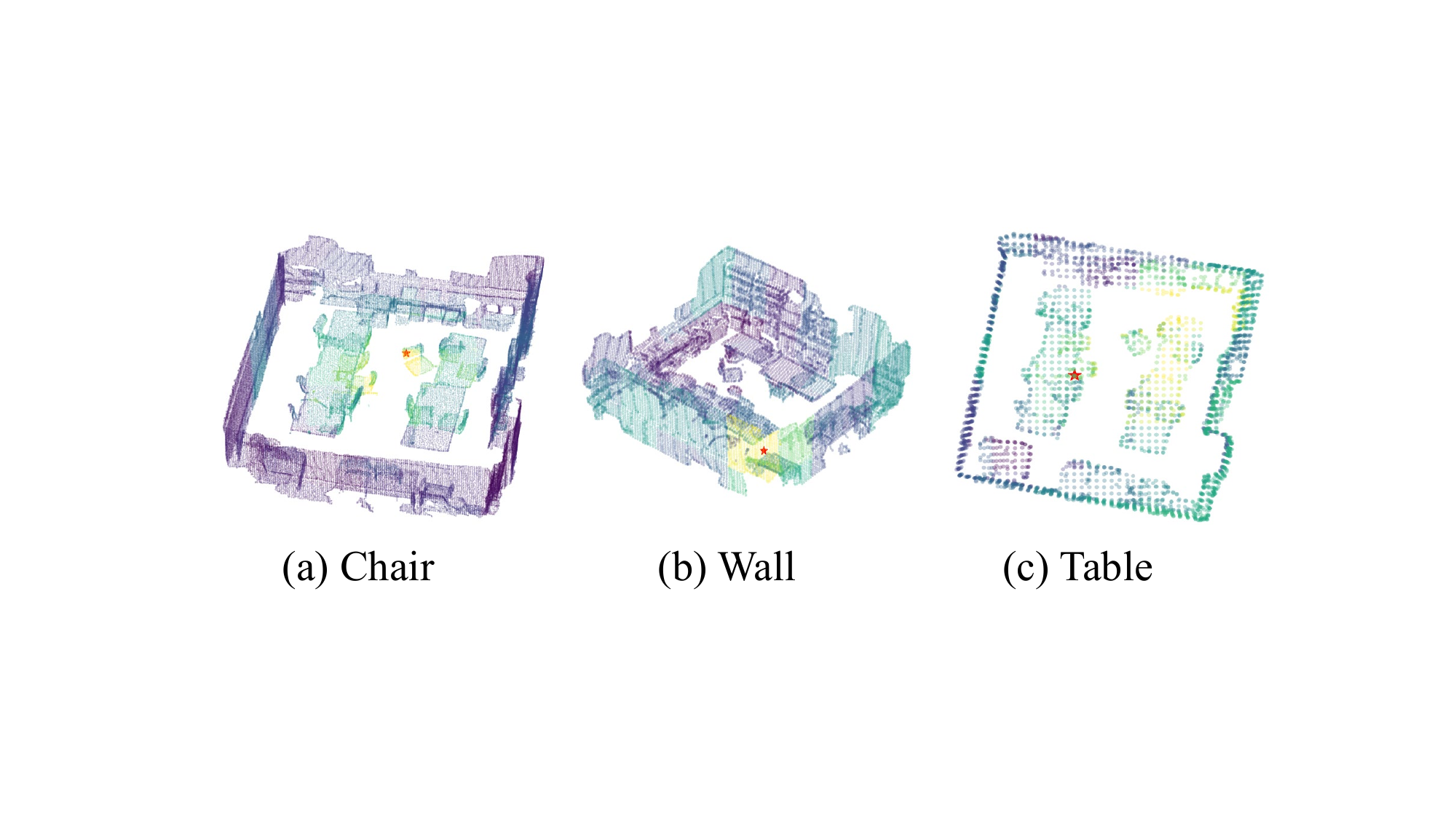}
    \vspace{-5mm}
    \caption{Visualization of point-to-point attention map of SPA of different objects and stages. The red star represents the current point. Specifically, (a), (b), and (c) illustrate the visualization results corresponding to 1-st, 2-rd, and 3-th SP$^2$TLayer.}
    \vspace{-5mm}
    \label{fig:glb-and-cls-weight}
\end{figure}

By processing the attention maps from point-to-proxy, proxy-to-proxy, and proxy-to-point, Fig.~\ref{fig:glb-and-cls-weight} illustrates the point-to-point SPA attention map at different stages(with different downsampling rates). Specifically, Fig.~\ref{fig:glb-and-cls-weight}~(a), (b), and (c) illustrate the visualization results corresponding to 1-st, 2-rd, and 3-th SP$^2$TLayer, respectively. It is noted that both object semantics and TRB together influence the model's attention. Initially, in the early stage of the model, TRB primarily directs the attention map by emphasizing the spatial pattern. Object semantics increasingly dominate the model's attention as the model progresses to a deeper stage.

\section{Conclusion}
We present a novel method called Sparse Proxy Point Transformer (SP$^2$T) designed for 3D understanding. It incorporates a local proxy-based dual-stream transformer and uses sparse proxy attention to enable a global receptive field. Key innovations include a spatial-wise sample method for robust sampling, sparse proxy attention for efficient proxy-point feature interaction, and a dual-stream structure to balance local and global information processing. SP$^2$T demonstrates state-of-the-art performance across various tasks despite being improvable in terms of speed. Future improvements focus on parameter tuning and optimizing dual-stream dimensional balance.


\section*{Acknowledgments}
This work was supported by the National Natural Science Foundation of China (62433003) and the National Natural Science Foundation of China (62476017).

{
    \small
    \bibliographystyle{ieeenat_fullname}
    \bibliography{references}

\begin{thebibliography}{69}
\providecommand{\natexlab}[1]{#1}
\providecommand{\url}[1]{\texttt{#1}}
\expandafter\ifx\csname urlstyle\endcsname\relax
  \providecommand{\doi}[1]{doi: #1}\else
  \providecommand{\doi}{doi: \begingroup \urlstyle{rm}\Url}\fi

\bibitem[Armeni et~al.(2016)Armeni, Sener, Zamir, Jiang, Brilakis, Fischer, and Savarese]{armeni20163d}
Iro Armeni, Ozan Sener, Amir~R Zamir, Helen Jiang, Ioannis Brilakis, Martin Fischer, and Silvio Savarese.
\newblock 3d semantic parsing of large-scale indoor spaces.
\newblock In \emph{Proceedings of the IEEE conference on computer vision and pattern recognition}, pages 1534--1543, 2016.

\bibitem[Behley et~al.(2019)Behley, Garbade, Milioto, Quenzel, Behnke, Stachniss, and Gall]{behley2019semantickitti}
Jens Behley, Martin Garbade, Andres Milioto, Jan Quenzel, Sven Behnke, Cyrill Stachniss, and Jurgen Gall.
\newblock Semantickitti: A dataset for semantic scene understanding of lidar sequences.
\newblock In \emph{Proceedings of the IEEE/CVF international conference on computer vision}, pages 9297--9307, 2019.

\bibitem[Berman et~al.(2018)Berman, Triki, and Blaschko]{berman2018lovasz}
Maxim Berman, Amal~Rannen Triki, and Matthew~B Blaschko.
\newblock The lov{\'a}sz-softmax loss: A tractable surrogate for the optimization of the intersection-over-union measure in neural networks.
\newblock In \emph{Proceedings of the IEEE conference on computer vision and pattern recognition}, pages 4413--4421, 2018.

\bibitem[Caesar et~al.(2020)Caesar, Bankiti, Lang, Vora, Liong, Xu, Krishnan, Pan, Baldan, and Beijbom]{caesar2020nuscenes}
Holger Caesar, Varun Bankiti, Alex~H Lang, Sourabh Vora, Venice~Erin Liong, Qiang Xu, Anush Krishnan, Yu Pan, Giancarlo Baldan, and Oscar Beijbom.
\newblock nuscenes: A multimodal dataset for autonomous driving.
\newblock In \emph{Proceedings of the IEEE/CVF conference on computer vision and pattern recognition}, pages 11621--11631, 2020.

\bibitem[Chen et~al.(2017)Chen, Ma, Wan, Li, and Xia]{chen2017multi}
Xiaozhi Chen, Huimin Ma, Ji Wan, Bo Li, and Tian Xia.
\newblock Multi-view 3d object detection network for autonomous driving.
\newblock In \emph{Proceedings of the IEEE conference on Computer Vision and Pattern Recognition}, pages 1907--1915, 2017.

\bibitem[Chen et~al.(2022)Chen, Dai, Chen, Liu, Dong, Yuan, and Liu]{chen2022mobile}
Yinpeng Chen, Xiyang Dai, Dongdong Chen, Mengchen Liu, Xiaoyi Dong, Lu Yuan, and Zicheng Liu.
\newblock Mobile-former: Bridging mobilenet and transformer.
\newblock In \emph{Proceedings of the IEEE/CVF conference on computer vision and pattern recognition}, pages 5270--5279, 2022.

\bibitem[Chen et~al.(2023)Chen, Liu, Zhang, Qi, and Jia]{chen2023voxelnext}
Yukang Chen, Jianhui Liu, Xiangyu Zhang, Xiaojuan Qi, and Jiaya Jia.
\newblock Voxelnext: Fully sparse voxelnet for 3d object detection and tracking.
\newblock In \emph{Proceedings of the IEEE/CVF Conference on Computer Vision and Pattern Recognition}, pages 21674--21683, 2023.

\bibitem[Cheng et~al.(2021)Cheng, Razani, Taghavi, Li, and Liu]{Cheng2021af2s3net}
Ran Cheng, Ryan Razani, Ehsan Taghavi, Enxu Li, and Bingbing Liu.
\newblock 2-s3net: Attentive feature fusion with adaptive feature selection for sparse semantic segmentation network.
\newblock In \emph{Proceedings of the IEEE/CVF conference on computer vision and pattern recognition}, pages 12547--12556, 2021.

\bibitem[Choy et~al.(2019)Choy, Gwak, and Savarese]{choy20194d}
Christopher Choy, JunYoung Gwak, and Silvio Savarese.
\newblock 4d spatio-temporal convnets: Minkowski convolutional neural networks.
\newblock In \emph{Proceedings of the IEEE/CVF conference on computer vision and pattern recognition}, pages 3075--3084, 2019.

\bibitem[Contributors(2023)]{pointcept2023}
Pointcept Contributors.
\newblock Pointcept: A codebase for point cloud perception research.
\newblock \url{https://github.com/Pointcept/Pointcept}, 2023.

\bibitem[Dai et~al.(2017)Dai, Chang, Savva, Halber, Funkhouser, and Nie{\ss}ner]{dai2017scannet}
Angela Dai, Angel~X Chang, Manolis Savva, Maciej Halber, Thomas Funkhouser, and Matthias Nie{\ss}ner.
\newblock Scannet: Richly-annotated 3d reconstructions of indoor scenes.
\newblock In \emph{Proceedings of the IEEE conference on computer vision and pattern recognition}, pages 5828--5839, 2017.

\bibitem[Darcet et~al.(2023)Darcet, Oquab, Mairal, and Bojanowski]{darcet2023vision}
Timoth{\'e}e Darcet, Maxime Oquab, Julien Mairal, and Piotr Bojanowski.
\newblock Vision transformers need registers.
\newblock \emph{arXiv preprint arXiv:2309.16588}, 2023.

\bibitem[Fan et~al.(2022)Fan, Pang, Zhang, Wang, Zhao, Wang, Wang, and Zhang]{fan2022embracing}
Lue Fan, Ziqi Pang, Tianyuan Zhang, Yu-Xiong Wang, Hang Zhao, Feng Wang, Naiyan Wang, and Zhaoxiang Zhang.
\newblock Embracing single stride 3d object detector with sparse transformer.
\newblock In \emph{Proceedings of the IEEE/CVF conference on computer vision and pattern recognition}, pages 8458--8468, 2022.

\bibitem[Felzenszwalb and Huttenlocher(2004)]{felzenszwalb2004efficient}
Pedro~F Felzenszwalb and Daniel~P Huttenlocher.
\newblock Efficient graph-based image segmentation.
\newblock \emph{International journal of computer vision}, 59:\penalty0 167--181, 2004.

\bibitem[Han et~al.(2023)Han, Ye, Han, Xia, Pan, Wan, Song, and Huang]{han2023agent}
Dongchen Han, Tianzhu Ye, Yizeng Han, Zhuofan Xia, Siyuan Pan, Pengfei Wan, Shiji Song, and Gao Huang.
\newblock Agent attention: On the integration of softmax and linear attention.
\newblock \emph{arXiv preprint arXiv:2312.08874}, 2023.

\bibitem[Han et~al.(2022)Han, Wang, Chen, Chen, Guo, Liu, Tang, Xiao, Xu, Xu, et~al.]{han2022survey}
Kai Han, Yunhe Wang, Hanting Chen, Xinghao Chen, Jianyuan Guo, Zhenhua Liu, Yehui Tang, An Xiao, Chunjing Xu, Yixing Xu, et~al.
\newblock A survey on vision transformer.
\newblock \emph{IEEE transactions on pattern analysis and machine intelligence}, 45\penalty0 (1):\penalty0 87--110, 2022.

\bibitem[Hao et~al.(2024)Hao, Li, Zhang, Li, Yin, Jung, Park, Yoo, Zhao, and Zhang]{hao2024mapdistill}
Xiaoshuai Hao, Ruikai Li, Hui Zhang, Dingzhe Li, Rong Yin, Sangil Jung, Seung-In Park, ByungIn Yoo, Haimei Zhao, and Jing Zhang.
\newblock Mapdistill: Boosting efficient camera-based hd map construction via camera-lidar fusion model distillation.
\newblock In \emph{European Conference on Computer Vision}, pages 166--183. Springer, 2024.

\bibitem[He et~al.(2024)He, Li, Zhang, and Zhang]{he2024scatterformer}
Chenhang He, Ruihuang Li, Guowen Zhang, and Lei Zhang.
\newblock Scatterformer: Efficient voxel transformer with scattered linear attention.
\newblock \emph{arXiv preprint arXiv:2401.00912}, 2024.

\bibitem[Huang et~al.(2023)Huang, Zhou, Cao, He, and Tan]{huang2023vision}
Huaibo Huang, Xiaoqiang Zhou, Jie Cao, Ran He, and Tieniu Tan.
\newblock Vision transformer with super token sampling.
\newblock In \emph{Proceedings of the IEEE/CVF conference on computer vision and pattern recognition}, pages 22690--22699, 2023.

\bibitem[Jaegle et~al.(2021{\natexlab{a}})Jaegle, Borgeaud, Alayrac, Doersch, Ionescu, Ding, Koppula, Zoran, Brock, Shelhamer, et~al.]{jaegle2021perceiverio}
Andrew Jaegle, Sebastian Borgeaud, Jean-Baptiste Alayrac, Carl Doersch, Catalin Ionescu, David Ding, Skanda Koppula, Daniel Zoran, Andrew Brock, Evan Shelhamer, et~al.
\newblock Perceiver io: A general architecture for structured inputs \& outputs.
\newblock \emph{arXiv preprint arXiv:2107.14795}, 2021{\natexlab{a}}.

\bibitem[Jaegle et~al.(2021{\natexlab{b}})Jaegle, Gimeno, Brock, Vinyals, Zisserman, and Carreira]{jaegle2021perceiver}
Andrew Jaegle, Felix Gimeno, Andy Brock, Oriol Vinyals, Andrew Zisserman, and Joao Carreira.
\newblock Perceiver: General perception with iterative attention.
\newblock In \emph{International conference on machine learning}, pages 4651--4664. PMLR, 2021{\natexlab{b}}.

\bibitem[Jiang et~al.(2020)Jiang, Zhao, Shi, Liu, Fu, and Jia]{jiang2020pointgroup}
Li Jiang, Hengshuang Zhao, Shaoshuai Shi, Shu Liu, Chi-Wing Fu, and Jiaya Jia.
\newblock Pointgroup: Dual-set point grouping for 3d instance segmentation.
\newblock In \emph{Proceedings of the IEEE/CVF conference on computer vision and Pattern recognition}, pages 4867--4876, 2020.

\bibitem[Khan et~al.(2022)Khan, Naseer, Hayat, Zamir, Khan, and Shah]{khan2022transformers}
Salman Khan, Muzammal Naseer, Munawar Hayat, Syed~Waqas Zamir, Fahad~Shahbaz Khan, and Mubarak Shah.
\newblock Transformers in vision: A survey.
\newblock \emph{ACM computing surveys (CSUR)}, 54\penalty0 (10s):\penalty0 1--41, 2022.

\bibitem[Lai et~al.(2022)Lai, Liu, Jiang, Wang, Zhao, Liu, Qi, and Jia]{lai2022stratified}
Xin Lai, Jianhui Liu, Li Jiang, Liwei Wang, Hengshuang Zhao, Shu Liu, Xiaojuan Qi, and Jiaya Jia.
\newblock Stratified transformer for 3d point cloud segmentation.
\newblock In \emph{Proceedings of the IEEE/CVF conference on computer vision and pattern recognition}, pages 8500--8509, 2022.

\bibitem[Lai et~al.(2023)Lai, Chen, Lu, Liu, and Jia]{lai2023spherical}
Xin Lai, Yukang Chen, Fanbin Lu, Jianhui Liu, and Jiaya Jia.
\newblock Spherical transformer for lidar-based 3d recognition.
\newblock In \emph{Proceedings of the IEEE/CVF Conference on Computer Vision and Pattern Recognition}, pages 17545--17555, 2023.

\bibitem[Lang et~al.(2019)Lang, Vora, Caesar, Zhou, Yang, and Beijbom]{lang2019pointpillars}
Alex~H Lang, Sourabh Vora, Holger Caesar, Lubing Zhou, Jiong Yang, and Oscar Beijbom.
\newblock Pointpillars: Fast encoders for object detection from point clouds.
\newblock In \emph{Proceedings of the IEEE/CVF conference on computer vision and pattern recognition}, pages 12697--12705, 2019.

\bibitem[Lee et~al.(2019)Lee, Lee, Kim, Kosiorek, Choi, and Teh]{lee2019set}
Juho Lee, Yoonho Lee, Jungtaek Kim, Adam Kosiorek, Seungjin Choi, and Yee~Whye Teh.
\newblock Set transformer: A framework for attention-based permutation-invariant neural networks.
\newblock In \emph{International conference on machine learning}, pages 3744--3753. PMLR, 2019.

\bibitem[Li et~al.(2016)Li, Zhang, and Xia]{li2016vehicle}
Bo Li, Tianlei Zhang, and Tian Xia.
\newblock Vehicle detection from 3d lidar using fully convolutional network.
\newblock \emph{arXiv preprint arXiv:1608.07916}, 2016.

\bibitem[Li et~al.(2024)Li, Shan, Jiang, Xiao, Chang, He, Yu, and Ren]{li2024mlp}
Ruikai Li, Hao Shan, Han Jiang, Jianru Xiao, Yizhuo Chang, Yifan He, Haiyang Yu, and Yilong Ren.
\newblock E-mlp: Effortless online hd map construction with linear priors.
\newblock In \emph{2024 IEEE Intelligent Vehicles Symposium (IV)}, pages 1008--1014. IEEE, 2024.

\bibitem[Lin et~al.(2023)Lin, Zheng, Li, Chao, Wang, Wang, Tian, and Ji]{lin2023meta}
Haojia Lin, Xiawu Zheng, Lijiang Li, Fei Chao, Shanshan Wang, Yan Wang, Yonghong Tian, and Rongrong Ji.
\newblock Meta architecture for point cloud analysis.
\newblock In \emph{Proceedings of the IEEE/CVF Conference on Computer Vision and Pattern Recognition}, pages 17682--17691, 2023.

\bibitem[Liu et~al.(2021)Liu, Lin, Cao, Hu, Wei, Zhang, Lin, and Guo]{liu2021swin}
Ze Liu, Yutong Lin, Yue Cao, Han Hu, Yixuan Wei, Zheng Zhang, Stephen Lin, and Baining Guo.
\newblock Swin transformer: Hierarchical vision transformer using shifted windows.
\newblock In \emph{Proceedings of the IEEE/CVF international conference on computer vision}, pages 10012--10022, 2021.

\bibitem[Liu et~al.(2022)Liu, Hu, Lin, Yao, Xie, Wei, Ning, Cao, Zhang, Dong, et~al.]{liu2022swin}
Ze Liu, Han Hu, Yutong Lin, Zhuliang Yao, Zhenda Xie, Yixuan Wei, Jia Ning, Yue Cao, Zheng Zhang, Li Dong, et~al.
\newblock Swin transformer v2: Scaling up capacity and resolution.
\newblock In \emph{Proceedings of the IEEE/CVF conference on computer vision and pattern recognition}, pages 12009--12019, 2022.

\bibitem[Lyu et~al.(2023)Lyu, Zhang, Li, Liu, Yang, and Yuan]{lyu2023uedg}
Yixuan Lyu, Hong Zhang, Yan Li, Hanyang Liu, Yifan Yang, and Ding Yuan.
\newblock Uedg: uncertainty-edge dual guided camouflage object detection.
\newblock \emph{IEEE Transactions on Multimedia}, 26:\penalty0 4050--4060, 2023.

\bibitem[Mao et~al.(2021)Mao, Zhang, Zheng, Ma, Peng, Ding, Zhang, Han, et~al.]{mao2021dual}
Mingyuan Mao, Renrui Zhang, Honghui Zheng, Teli Ma, Yan Peng, Errui Ding, Baochang Zhang, Shumin Han, et~al.
\newblock Dual-stream network for visual recognition.
\newblock \emph{Advances in Neural Information Processing Systems}, 34:\penalty0 25346--25358, 2021.

\bibitem[Maturana and Scherer(2015)]{maturana2015voxnet}
Daniel Maturana and Sebastian Scherer.
\newblock Voxnet: A 3d convolutional neural network for real-time object recognition.
\newblock In \emph{2015 IEEE/RSJ international conference on intelligent robots and systems (IROS)}, pages 922--928. IEEE, 2015.

\bibitem[Nekrasov et~al.(2021)Nekrasov, Schult, Litany, Leibe, and Engelmann]{nekrasov2021mix3d}
Alexey Nekrasov, Jonas Schult, Or Litany, Bastian Leibe, and Francis Engelmann.
\newblock Mix3d: Out-of-context data augmentation for 3d scenes.
\newblock In \emph{2021 international conference on 3d vision (3dv)}, pages 116--125. IEEE, 2021.

\bibitem[Park et~al.(2022)Park, Jeong, Cho, and Park]{park2022fast}
Chunghyun Park, Yoonwoo Jeong, Minsu Cho, and Jaesik Park.
\newblock Fast point transformer.
\newblock In \emph{Proceedings of the IEEE/CVF conference on computer vision and pattern recognition}, pages 16949--16958, 2022.

\bibitem[Park et~al.(2023)Park, Lee, Kim, Xiong, and Kim]{park2023self}
Jinyoung Park, Sanghyeok Lee, Sihyeon Kim, Yunyang Xiong, and Hyunwoo~J Kim.
\newblock Self-positioning point-based transformer for point cloud understanding.
\newblock In \emph{Proceedings of the IEEE/CVF conference on computer vision and pattern recognition}, pages 21814--21823, 2023.

\bibitem[Park and Kim(2022)]{park2022vision}
Namuk Park and Songkuk Kim.
\newblock How do vision transformers work?
\newblock \emph{arXiv preprint arXiv:2202.06709}, 2022.

\bibitem[Qi et~al.(2017{\natexlab{a}})Qi, Su, Mo, and Guibas]{qi2017pointnet}
Charles~R Qi, Hao Su, Kaichun Mo, and Leonidas~J Guibas.
\newblock Pointnet: Deep learning on point sets for 3d classification and segmentation.
\newblock In \emph{Proceedings of the IEEE conference on computer vision and pattern recognition}, pages 652--660, 2017{\natexlab{a}}.

\bibitem[Qi et~al.(2017{\natexlab{b}})Qi, Yi, Su, and Guibas]{qi2017pointnet++}
Charles~Ruizhongtai Qi, Li Yi, Hao Su, and Leonidas~J Guibas.
\newblock Pointnet++: Deep hierarchical feature learning on point sets in a metric space.
\newblock \emph{Advances in neural information processing systems}, 30, 2017{\natexlab{b}}.

\bibitem[Qian et~al.(2022)Qian, Li, Peng, Mai, Hammoud, Elhoseiny, and Ghanem]{qian2022pointnext}
Guocheng Qian, Yuchen Li, Houwen Peng, Jinjie Mai, Hasan Hammoud, Mohamed Elhoseiny, and Bernard Ghanem.
\newblock Pointnext: Revisiting pointnet++ with improved training and scaling strategies.
\newblock \emph{Advances in neural information processing systems}, 35:\penalty0 23192--23204, 2022.

\bibitem[Rozenberszki et~al.(2022)Rozenberszki, Litany, and Dai]{rozenberszki2022language}
David Rozenberszki, Or Litany, and Angela Dai.
\newblock Language-grounded indoor 3d semantic segmentation in the wild.
\newblock In \emph{European Conference on Computer Vision}, pages 125--141. Springer, 2022.

\bibitem[Schult et~al.(2023)Schult, Engelmann, Hermans, Litany, Tang, and Leibe]{schult2023mask3d}
Jonas Schult, Francis Engelmann, Alexander Hermans, Or Litany, Siyu Tang, and Bastian Leibe.
\newblock Mask3d: Mask transformer for 3d semantic instance segmentation.
\newblock In \emph{2023 IEEE International Conference on Robotics and Automation (ICRA)}, pages 8216--8223. IEEE, 2023.

\bibitem[Song et~al.(2017)Song, Yu, Zeng, Chang, Savva, and Funkhouser]{song2017semantic}
Shuran Song, Fisher Yu, Andy Zeng, Angel~X Chang, Manolis Savva, and Thomas Funkhouser.
\newblock Semantic scene completion from a single depth image.
\newblock In \emph{Proceedings of the IEEE conference on computer vision and pattern recognition}, pages 1746--1754, 2017.

\bibitem[Srivastava et~al.(2014)Srivastava, Hinton, Krizhevsky, Sutskever, and Salakhutdinov]{srivastava2014dropout}
Nitish Srivastava, Geoffrey Hinton, Alex Krizhevsky, Ilya Sutskever, and Ruslan Salakhutdinov.
\newblock Dropout: a simple way to prevent neural networks from overfitting.
\newblock \emph{The journal of machine learning research}, 15\penalty0 (1):\penalty0 1929--1958, 2014.

\bibitem[Su et~al.(2015)Su, Maji, Kalogerakis, and Learned-Miller]{su2015multi}
Hang Su, Subhransu Maji, Evangelos Kalogerakis, and Erik Learned-Miller.
\newblock Multi-view convolutional neural networks for 3d shape recognition.
\newblock In \emph{Proceedings of the IEEE international conference on computer vision}, pages 945--953, 2015.

\bibitem[Sun et~al.(2020)Sun, Kretzschmar, Dotiwalla, Chouard, Patnaik, Tsui, Guo, Zhou, Chai, Caine, et~al.]{sun2020scalability}
Pei Sun, Henrik Kretzschmar, Xerxes Dotiwalla, Aurelien Chouard, Vijaysai Patnaik, Paul Tsui, James Guo, Yin Zhou, Yuning Chai, Benjamin Caine, et~al.
\newblock Scalability in perception for autonomous driving: Waymo open dataset.
\newblock In \emph{Proceedings of the IEEE/CVF conference on computer vision and pattern recognition}, pages 2446--2454, 2020.

\bibitem[Taunk et~al.(2019)Taunk, De, Verma, and Swetapadma]{taunk2019brief}
Kashvi Taunk, Sanjukta De, Srishti Verma, and Aleena Swetapadma.
\newblock A brief review of nearest neighbor algorithm for learning and classification.
\newblock In \emph{2019 international conference on intelligent computing and control systems (ICCS)}, pages 1255--1260. IEEE, 2019.

\bibitem[Vaswani(2017)]{vaswani2017attention}
A Vaswani.
\newblock Attention is all you need.
\newblock \emph{Advances in Neural Information Processing Systems}, 2017.

\bibitem[Wang et~al.(2023)Wang, Shi, Shi, Lei, Wang, He, Schiele, and Wang]{wang2023dsvt}
Haiyang Wang, Chen Shi, Shaoshuai Shi, Meng Lei, Sen Wang, Di He, Bernt Schiele, and Liwei Wang.
\newblock Dsvt: Dynamic sparse voxel transformer with rotated sets.
\newblock In \emph{Proceedings of the IEEE/CVF Conference on Computer Vision and Pattern Recognition}, pages 13520--13529, 2023.

\bibitem[Wang(2023)]{wang2023octformer}
Peng-Shuai Wang.
\newblock Octformer: Octree-based transformers for 3d point clouds.
\newblock \emph{ACM Transactions on Graphics (TOG)}, 42\penalty0 (4):\penalty0 1--11, 2023.

\bibitem[Wang et~al.(2020)Wang, Li, Khabsa, Fang, and Ma]{wang2020linformer}
Sinong Wang, Belinda~Z Li, Madian Khabsa, Han Fang, and Hao Ma.
\newblock Linformer: Self-attention with linear complexity.
\newblock \emph{arXiv preprint arXiv:2006.04768}, 2020.

\bibitem[Wang et~al.(2022)Wang, Guizilini, Zhang, Wang, Zhao, and Solomon]{wang2022detr3d}
Yue Wang, Vitor~Campagnolo Guizilini, Tianyuan Zhang, Yilun Wang, Hang Zhao, and Justin Solomon.
\newblock Detr3d: 3d object detection from multi-view images via 3d-to-2d queries.
\newblock In \emph{Conference on Robot Learning}, pages 180--191. PMLR, 2022.

\bibitem[Wu et~al.(2022)Wu, Lao, Jiang, Liu, and Zhao]{wu2022point}
Xiaoyang Wu, Yixing Lao, Li Jiang, Xihui Liu, and Hengshuang Zhao.
\newblock Point transformer v2: Grouped vector attention and partition-based pooling.
\newblock \emph{Advances in Neural Information Processing Systems}, 35:\penalty0 33330--33342, 2022.

\bibitem[Wu et~al.(2024)Wu, Jiang, Wang, Liu, Liu, Qiao, Ouyang, He, and Zhao]{wu2024point}
Xiaoyang Wu, Li Jiang, Peng-Shuai Wang, Zhijian Liu, Xihui Liu, Yu Qiao, Wanli Ouyang, Tong He, and Hengshuang Zhao.
\newblock Point transformer v3: Simpler faster stronger.
\newblock In \emph{Proceedings of the IEEE/CVF Conference on Computer Vision and Pattern Recognition}, pages 4840--4851, 2024.

\bibitem[Yan et~al.(2020)Yan, Zheng, Li, Wang, and Cui]{yan2020pointasnl}
Xu Yan, Chaoda Zheng, Zhen Li, Sheng Wang, and Shuguang Cui.
\newblock Pointasnl: Robust point clouds processing using nonlocal neural networks with adaptive sampling.
\newblock In \emph{Proceedings of the IEEE/CVF conference on computer vision and pattern recognition}, pages 5589--5598, 2020.

\bibitem[Yan et~al.(2022)Yan, Gao, Zheng, Zheng, Zhang, Cui, and Li]{yan20222dpass}
Xu Yan, Jiantao Gao, Chaoda Zheng, Chao Zheng, Ruimao Zhang, Shuguang Cui, and Zhen Li.
\newblock 2dpass: 2d priors assisted semantic segmentation on lidar point clouds.
\newblock In \emph{European Conference on Computer Vision}, pages 677--695. Springer, 2022.

\bibitem[Yang et~al.(2023)Yang, Guo, Xiong, Liu, Pan, Wang, Tong, and Guo]{yang2023swin3d}
Yu-Qi Yang, Yu-Xiao Guo, Jian-Yu Xiong, Yang Liu, Hao Pan, Peng-Shuai Wang, Xin Tong, and Baining Guo.
\newblock Swin3d: A pretrained transformer backbone for 3d indoor scene understanding.
\newblock \emph{arXiv preprint arXiv:2304.06906}, 2023.

\bibitem[Yin et~al.(2021)Yin, Zhou, and Krahenbuhl]{yin2021center}
Tianwei Yin, Xingyi Zhou, and Philipp Krahenbuhl.
\newblock Center-based 3d object detection and tracking.
\newblock In \emph{Proceedings of the IEEE/CVF conference on computer vision and pattern recognition}, pages 11784--11793, 2021.

\bibitem[Zhang et~al.(2024{\natexlab{a}})Zhang, Chen, Gao, Li, Liu, and Hu]{zhang2024safdnet}
Gang Zhang, Junnan Chen, Guohuan Gao, Jianmin Li, Si Liu, and Xiaolin Hu.
\newblock Safdnet: A simple and effective network for fully sparse 3d object detection.
\newblock In \emph{Proceedings of the IEEE/CVF Conference on Computer Vision and Pattern Recognition}, pages 14477--14486, 2024{\natexlab{a}}.

\bibitem[Zhang et~al.(2024{\natexlab{b}})Zhang, Wan, He, Song, Yang, and Yuan]{zhang2024sparse}
Hong Zhang, Jiaxu Wan, Ziqi He, Jianbo Song, Yifan Yang, and Ding Yuan.
\newblock Sparse agent transformer for unified voxel and image feature extraction and fusion.
\newblock \emph{Information Fusion}, page 102455, 2024{\natexlab{b}}.

\bibitem[Zhang et~al.(2024{\natexlab{c}})Zhang, Wan, Zhang, Yuan, Li, and Yang]{zhang2024p2ftrack}
Hong Zhang, Jiaxu Wan, Jing Zhang, Ding Yuan, XuLiang Li, and Yifan Yang.
\newblock P2ftrack: Multi-object tracking with motion prior and feature posterior.
\newblock \emph{ACM Transactions on Multimedia Computing, Communications and Applications}, 2024{\natexlab{c}}.

\bibitem[Zhang et~al.(2025{\natexlab{a}})Zhang, Lyu, He, Li, Li, Yuan, and Yang]{zhang2025coddiff}
Hong Zhang, Yixuan Lyu, Tian He, Xuliang Li, Yawei Li, Ding Yuan, and Yifan Yang.
\newblock Coddiff: Prior leading diffusion model for camouflage object detection.
\newblock \emph{Knowledge-Based Systems}, page 113381, 2025{\natexlab{a}}.

\bibitem[Zhang et~al.(2025{\natexlab{b}})Zhang, Song, Liu, Han, Yang, and Ma]{zhang2025awaretrack}
Hong Zhang, Jianbo Song, Hanyang Liu, Yang Han, Yifan Yang, and Huimin Ma.
\newblock Awaretrack: Object awareness for visual tracking via templates interaction.
\newblock \emph{Image and Vision Computing}, 154:\penalty0 105363, 2025{\natexlab{b}}.

\bibitem[Zhang et~al.(2021)Zhang, Zeng, Guo, Gao, Fu, and Shi]{zhang2021dspoint}
Renrui Zhang, Ziyao Zeng, Ziyu Guo, Xinben Gao, Kexue Fu, and Jianbo Shi.
\newblock Dspoint: Dual-scale point cloud recognition with high-frequency fusion.
\newblock \emph{arXiv preprint arXiv:2111.10332}, 2021.

\bibitem[Zhao et~al.(2019)Zhao, Jiang, Fu, and Jia]{zhao2019pointweb}
Hengshuang Zhao, Li Jiang, Chi-Wing Fu, and Jiaya Jia.
\newblock Pointweb: Enhancing local neighborhood features for point cloud processing.
\newblock In \emph{Proceedings of the IEEE/CVF conference on computer vision and pattern recognition}, pages 5565--5573, 2019.

\bibitem[Zhao et~al.(2021)Zhao, Jiang, Jia, Torr, and Koltun]{zhao2021point}
Hengshuang Zhao, Li Jiang, Jiaya Jia, Philip~HS Torr, and Vladlen Koltun.
\newblock Point transformer.
\newblock In \emph{Proceedings of the IEEE/CVF international conference on computer vision}, pages 16259--16268, 2021.

\bibitem[Zhou and Tuzel(2018)]{zhou2018voxelnet}
Yin Zhou and Oncel Tuzel.
\newblock Voxelnet: End-to-end learning for point cloud based 3d object detection.
\newblock In \emph{Proceedings of the IEEE conference on computer vision and pattern recognition}, pages 4490--4499, 2018.

\end{thebibliography}
}

\clearpage
\appendix

\vspace{-4mm}
\section{Experiment and Discussion}

\begin{table}[t]
        \begin{minipage}{\linewidth}
            \centering
                \tablestyle{10pt}{1.1}
                \begin{tabular}{lcccc}
\toprule
Indoor Sem. Seg.&\multicolumn{2}{c}{ScanNet~\cite{dai2017scannet}} &\multicolumn{2}{c}{ScanNet200~\cite{rozenberszki2022language}}  \\\cmidrule(lr){1-1} \cmidrule(lr){2-3} \cmidrule(lr){4-5} 
Methods with same TTA &Val &Test &Val &Test \\\midrule
 MinkUnet~\cite{choy20194d} &72.2 &73.4 &25.0 &25.3 \\
 OctFormer~\cite{wang2023octformer} (Rep.) & 74.6 & 70.7 & 31.9 & 31.0  \\
  OctFormer~\cite{wang2023octformer} (Off.) & 75.7 & - & 32.6 &  - \\
Swin3D~\cite{yang2023swin3d} (Off.) & 76.4 & - & - & -  \\
 Swin3D~\cite{yang2023swin3d} (Rep.) & 76.6 & 71.4 & - & -  \\
 PTv3~\cite{wu2024point} (Off.) & 77.5 & 73.6 & 35.2 & 34.0  \\
 \rowcolor{gray!20}
 SP$^2$T &\textbf{78.7} & \textbf{74.9} & \textbf{37.0} & \textbf{35.2}   \\
\bottomrule
\end{tabular}

                \vspace{-2.5mm}
                \caption{Indoor instance segmentation with same TTA between Val and Test set. Rep. means the model uses the code of Pointcept and reproduces it by ours. Off. means the model uses official weight and code.}\label{tab:appendix_indor_sem_real}
    \end{minipage} \\
    \vspace{-2mm}
\end{table}

        

\definecolor{vscgreen}{RGB}{106,200,85} 

\begin{algorithm}[t]
\caption{\footnotesize Pytorch-Style Pesade-code of Spatial-wise Sampling}
\label{alg:grid-search}
\lstset{
  language=Python,
  basicstyle=\ttfamily\fontsize{7pt}{10pt}\selectfont,
  commentstyle=\color{vscgreen}, 
}
\begin{lstlisting}
def Spatial_Wise_Sampling(
    s_min: float, s_max: float,    # min/max cell size
    cnt_low: int, cnt_high: int,   # target count range
    max_iter: int,                 # max iterations
    grid_range: float              # AABB size
) -> float:                        # optimal grid size
    l, r = s_min, s_max
    for _ in range(max_iter):
        grid_size = (l + r) / 2
        grid_shape = ceil(grid_range / grid_size)
        cell_count = prod(grid_shape)
        
        if cnt_low <= cell_count <= cnt_high:
            return grid_size
        elif cell_count < cnt_low:
            r = grid_size          # too sparse
        else:
            l = grid_size          # too dense
    return (l + r) / 2             # fallback
\end{lstlisting}
\end{algorithm}


\vspace{-2mm}
\subsection{ScanNet Test Set}

According to~\cite{schult2023mask3d, yang2023swin3d, wu2024point}, there is a significant test time augmentation (TTA) difference between the val and test sets of Scannet~\cite{dai2017scannet} and Scannet200~\cite{rozenberszki2022language}. The additional TTA includes incorporating data from the validation set for training, combining results from multiple training models~\cite{lai2022stratified}, and applying over-segmentation~\cite{felzenszwalb2004efficient}.

For a fair comparison, we evaluated some SOTA models~\cite{choy20194d, wang2023octformer, yang2023swin3d, wu2024point} with an official or reproduction weights file using Val's TTA, as shown in Fig.~\ref{tab:appendix_indor_sem_real}. The experimental findings indicate that, without employing additional TTA, the result in the test set for the majority models~\cite{wang2023octformer, yang2023swin3d, wu2024point} tends to be less than those on the validation set, rather than exceeding them. And it can be found that our model archive the best result both in the val and test set of Scannet and Scannet200.

We are still working on over-segmentation and may update our model's test results employing over-segmentation in the final version of the paper. Furthermore, we recommend that future research ensure consistency in the TTA between the validation and test sets or at least make the TTA on the test set openly available. In 3D understanding, the focus should be on improving network design and training methodologies rather than using more testing tricks.

\subsection{Pesade-code of Spatial-wise Sampling}
The pesade-code of spatial-wise sampling is shown in Alg.~\ref{alg:grid-search}. Spatial-wise sampling efficiently discerns the ideal proxy spacing by considering the AABB sizes of various points. The method is designed to maintain the proxies count within the bounds of $N_{max}$ and $N_{min}$, using a bisection approach to determine the optimal proxy spacing $L_p$. If the number of proxies exceeds $N_{max}$, the proxy spacing $L_p$ is reduced, and the total number of proxies is recalculated.

\vspace{-1mm}
\subsection{Visualization}

\begin{figure}[t]
    \centering
    \includegraphics[width=\linewidth]{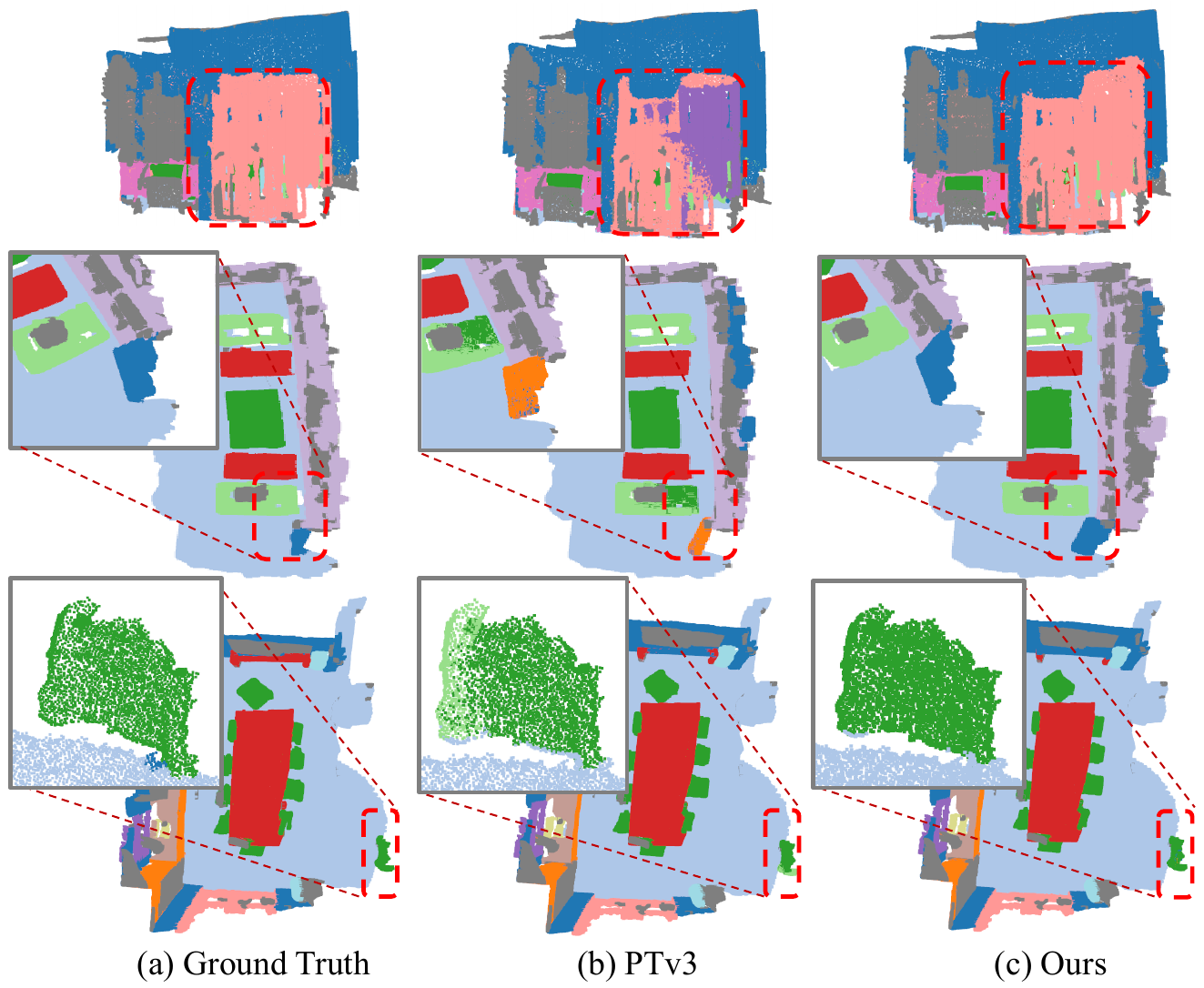}
    \vspace{-5mm}
    \caption{Result Comparison in ScanNet.}
    \vspace{-4mm}
    \label{fig:compare}
\end{figure}

\begin{figure}[t]
    \centering
    \includegraphics[width=\linewidth]{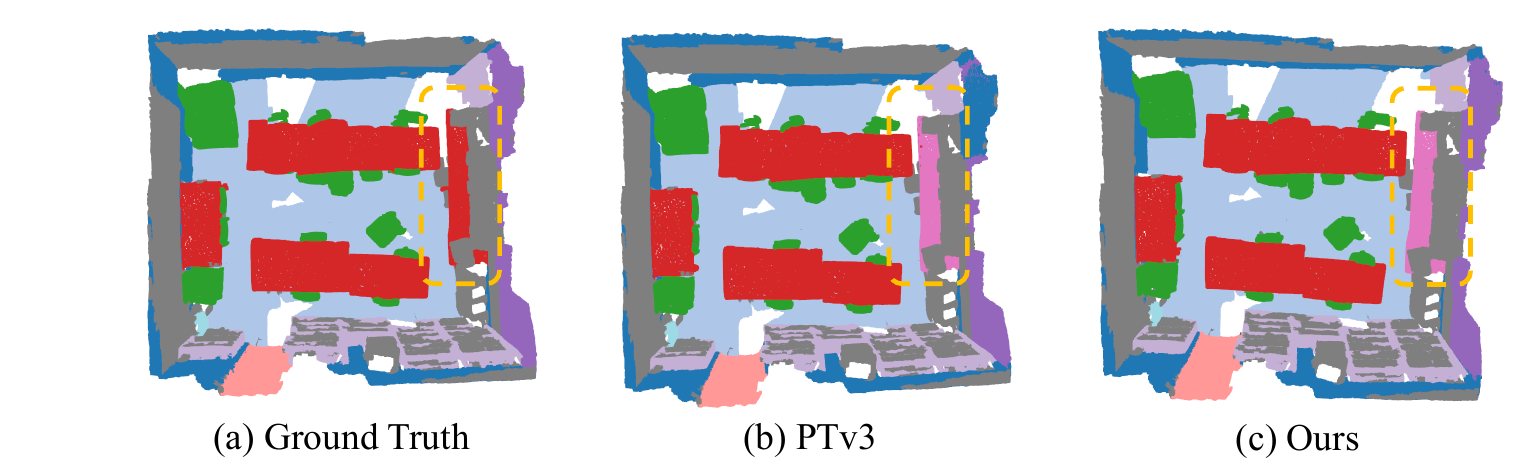}
    \vspace{-5mm}
    \caption{Failed cases in ScanNet.}
    \vspace{-6mm}
    \label{fig:failed}
\end{figure}

\nbf{Result Comparison.}
Fig.~\ref{fig:compare} compares the visualization of our method and PTv3~\cite{wu2024point} on the Scannet dataset~\cite{dai2017scannet}. Specifically, Fig.~\ref{fig:compare}~(a) presents the ground truth, Fig.~\ref{fig:compare}~(b) depicts the result from PTv3, and Fig.~\ref{fig:compare}~(c) illustrates our result. The visualizations indicate that, due to the global receptive field provided by the proxy, our method achieves more consistent and dependable segmentation results overall, thus enhancing segmentation performance.

\nbf{Failed Cases.} 
Fig.~\ref{fig:failed} illustrates the failed cases of our method on the Scannet dataset~\cite{dai2017scannet}. In Fig.~\ref{fig:failed}, it is apparent that the proxy has failed to address the classification error attributed to location fusion. Consequently, achieving a proper balance between local and global information still requires further investigation.

\begin{figure*}[t]
    \centering
    \includegraphics[width=\linewidth]{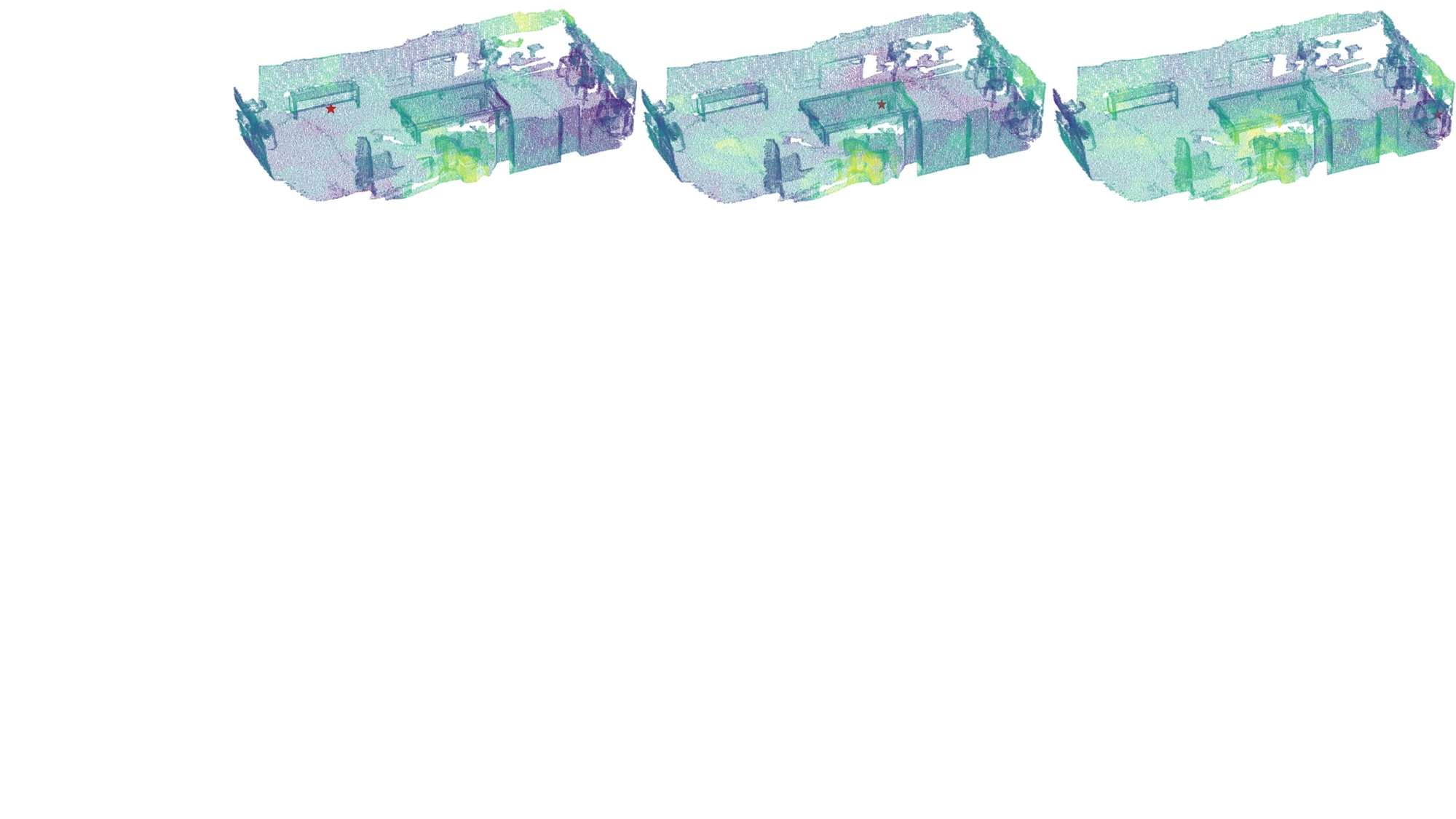}
    \vspace{-5mm}
    \caption{Visualization of the point-point attention map under FPS-based sampling. The red star represents the current point.}
    \vspace{-3mm}
    \label{fig:glb-attn-weights-cmp}
\end{figure*}

\nbf{Over-fitting due to FPS-based Sampling.}
Within the ablation study, the experiments show that FPS-based sampling performs poorly compared to alternative sampling methods. We visualized the point-proxy attention maps for FPS-based sampling methods to investigate this further, as illustrated in Fig.~\ref{fig:glb-attn-weights-cmp}. The visualization demonstrates that the attention map resulting from FPS-based sampling exhibits static and repetitive patterns, with its attention not influenced by the proxy's location. Consequently, we contend that the sampling method based on FPS leads to significant overfitting of the model because of the scene leakage from FPS.

\begin{table}[t]
\centering
\tablestyle{7.5pt}{1.1}
\begin{tabular}{cc|ccc|c}
    \toprule
    TRB & Share TRB & mIoU & mAcc & allAcc& Time    \\
    \midrule
    
    \usym{2714} & \usym{2717}          & 78.33   & 86.17 & 92.36  & 81ms \\
    \rowcolor{gray!20}
    \usym{2714} & \usym{2714}          & \textbf{78.71}   & \textbf{86.23} & \textbf{92.51}     & \textbf{74ms}  \\
    \bottomrule
\end{tabular}
\vspace{-2mm}
\caption{Ablation study about sharing of TRB.}
\label{tab:param-share}
\vspace{-2mm}
\end{table}

\begin{figure}[t]
    \centering
    \includegraphics[width=0.9\linewidth]{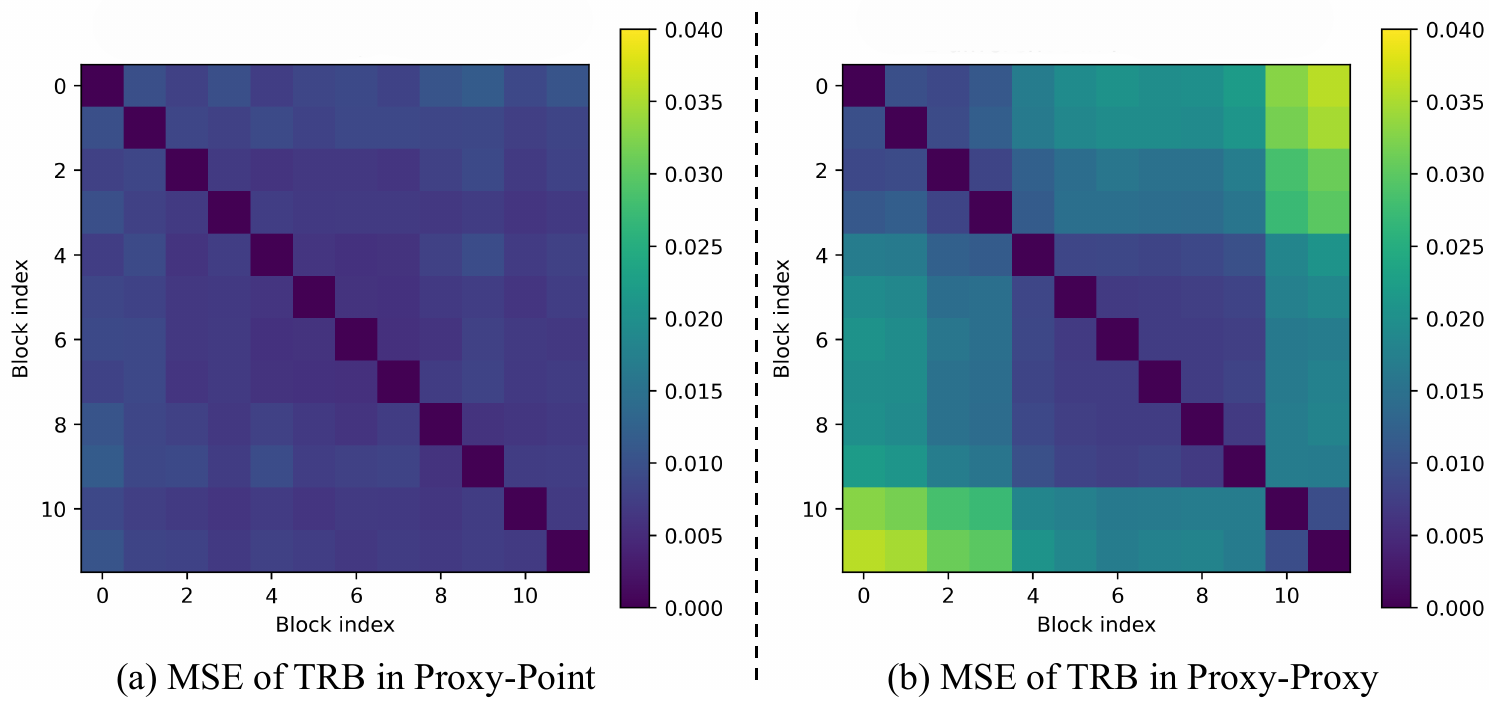}
    \vspace{-3mm}
    \caption{(a) MSE of TRB in different proxy-point interaction layers. (b) MSE of TRB in different proxy-proxy interaction layers.}
    \vspace{-6mm}
    \label{fig:rpe-diff}
\end{figure}

\subsection{Shared Table Relative Bias}
During the implementation phase, the point and proxy positions remain unchanged within the same layer. This means that for each instance of sparse attention in this layer, the relative position input provided to the relative position encoding module remains constant. Consequently, it is possible to compute the relative bias for a layer with a single invocation. Building on this optimization, we share all relative bias values across each layer, thus reducing model complexity and computational demand. Additionally, since the proxy positions are constant throughout the network, we test sharing the relative bias amongst proxies across the entire network and different layers.

Tab.~\ref{tab:param-share} compares the accuracy of the model w/ and w/o the shared TRB. The experiment shows that the shared TRB improves the model's accuracy and reduces the inference time. Furthermore, Fig.~\ref{fig:rpe-diff} shows the MSE distance for TRB during point-proxy and proxy-proxy interaction in different layers. TRB demonstrates stage-level similarity in the proxy-proxy interaction, while all TRB is similar in the point-proxy interaction. Consequently, the sharing of TRB improves model accuracy and reduces inference time.

\begin{table}[t]
    \centering
        \tablestyle{4pt}{1.1}
        \begin{tabular}{l|cccc}\toprule
Efficiency & \multicolumn{2}{c}{Indoor (ScanNet~\cite{dai2017scannet})} & \multicolumn{2}{c}{Outdoor (nuScenes~\cite{caesar2020nuscenes})}\\ \cmidrule(lr){1-1} \cmidrule(lr){2-3} \cmidrule(lr){4-5}
Methods & mIOU &Latency  & mIOU &Latency  \\
\midrule
MinkUNet~\cite{choy20194d} & 72.2 & 90ms  & 73.3 & 48ms  \\
PTv2~\cite{wu2022point} & 75.4 & 191ms  & 80.2 & 146ms  \\
PTv3~\cite{wu2024point}  & 77.5 & \textbf{61ms} & 80.4 & \textbf{44ms}  \\
PTv3$^\dag$~\cite{wu2024point} & 77.6 & 73ms & 80.5 & 53ms \\
\rowcolor{gray!20}
SP$^2$T & \textbf{78.7} & 74ms  & \textbf{81.2} & 54ms \\
\bottomrule
\end{tabular}
        \vspace{-2mm}
        \caption{Ablation study about efficiency of SP$^2$T. PTv3$^\dag$ refers to PTv3 with an increased number of channels to maintain the same latency as SP$^2$T.}\label{tab:abletion_model_efficiency}
        \vspace{-6mm}
\end{table}

\subsection{Model efficiency}
Tab.~\ref{tab:abletion_model_efficiency} presents the model's performance based on accuracy and latency for both the indoor (ScanNet) and outdoor (nuScenes) datasets, tested on a single RTX 4090. It is important to note that scaling the PTv3 model to match the size of SP2T does not substantially improve the metrics for either dataset. Our model achieves an ideal compromise between accuracy and speed while maintaining consistent accuracy.

\section{Implementation Details}
Our implementation primarily utilizes Pointcept~\cite{pointcept2023}, a specialized codebase focusing on point cloud perception and representation learning. The details of our implementation are detailed in this section.

\subsection{Training Settings}

\begin{table}[!t]
    \begin{minipage}{\linewidth}
    \centering
        \tablestyle{6pt}{1.1}
\begin{tabular}{lclc}\toprule
\multicolumn{2}{c}{Indoor Semantic} &\multicolumn{2}{c}{Indoor Instance} \\
\cmidrule(lr){1-2} \cmidrule(lr){3-4}
Config &Value &Config &Value \\\midrule
framework & / & frame & PointGroup \\
optimizer &AdamW &optimizer &AdamW \\
scheduler &Cosine &scheduler &Cosine \\
criteria &CrossEntropy~(1) &criteria & / \\
&Lovasz~\cite{berman2018lovasz}~(1) & & \\
weight decay &5e-2 &weight decay &5e-2 \\
batch size &12 &batch size &12 \\
datasets &ScanNet / S3DIS &datasets &ScanNet \\\midrule
\multicolumn{4}{l}{First Stage:} \\
learning rate &5e-3 &learning rate &5e-3 \\
block lr scaler &0.1 &block lr scaler &0.1 \\
warmup epochs &40 &warmup iters &40 \\
epochs &800 &epochs &800 \\\midrule
\multicolumn{4}{l}{Second Stage:} \\
learning rate &2e-4 &learning rate &2e-4 \\
block lr scaler &1.0 &block lr scaler &1.0 \\
warmup epochs &20 &warmup iters &20 \\
epochs &400 &epochs &400 \\
\bottomrule
\end{tabular}

        \vspace{-3mm}
        \caption{Indoor semantic / instance segmentation settings.}\label{tab:indoor_sem_seg_settings}
        \vspace{-4mm}
    \end{minipage}
\end{table}

\nbf{Datasets and metrics.} 
The ScanNet dataset~\cite{dai2017scannet, rozenberszki2022language}, frequently utilized in indoor real-world down-stream tasks, contains 1,513 room scans derived from RGB-D frames, with 1,201 scenes designated for training and 312 reserved for validation. Each point was categorized into one of the 20 semantic labels in ScanNet~\cite{dai2017scannet} and 200 semantic labels in ScanNet200~\cite{rozenberszki2022language}. In contrast, the S3DIS dataset~\cite{armeni20163d} covers 271 rooms in six areas within three buildings with 13 categories. 

nuScenes~\cite{caesar2020nuscenes} consists of 40,157 annotated samples, each containing six monocular camera images that cover a 360-degree field of view and a 32-beam LiDAR. According to the specifications of nuScenes, the dataset comprises 1000 scenarios, 1.4M images, and 400K point clouds. The training set covers 700 scenarios, and the validation and test sets contain 150 scenarios each. SemanticKITTI~\cite{behley2019semantickitti} originates from the KITTI Vision Benchmark Suite and is comprised of 22 sequences, with 19 designated for training and the other 3 reserved for testing. Waymo~\cite{sun2020scalability} is a frequently utilized benchmark for outdoor 3D perception, comprising a total of 1,150 point cloud sequences (exceeding 200K frames). Each frame encompasses an extensive perception range of 150m × 150m.  

For segmentation metrics, we utilize the mean class-wise intersection over union (mIoU) as the principal metric in ScanNet, ScanNet200, and S3DIS. Furthermore, following previous work, area 5 in S3DIS is designated for testing with a 6-fold cross-validation. For dectection metircs, all results are assessed by the conventional protocol employing 3D mean Average Precision (mAP) and its weighted version based on heading accuracy (mAPH).

\begin{table}[!t]
\begin{minipage}{\linewidth}
    \centering
        \tablestyle{6.5pt}{1.1}
        \begin{tabular}{lclc}\toprule
\multicolumn{4}{c}{Outdoor Semantic}  \\
\cmidrule(lr){1-4} 
Config &Value & Config &Value \\\midrule
optimizer &AdamW  &  batch size & 12\\
scheduler &Cosine & weight decay &5e-3 \\
criteria &CrossEntropy~(1) & datasets & NuScenes \\
&Lovasz~\cite{berman2018lovasz}~(1)  & & Sem.KITTI \\
&  & & Waymo \\\midrule
\multicolumn{4}{l}{First Stage:} \\
learning rate &2e-3   & epochs &50\\
block lr scaler & 1e-1 & warmup epochs &2 \\\midrule
\multicolumn{4}{l}{Second Stage:} \\
learning rate &2e-4   & epochs &30\\
block lr scaler & 1.0 & warmup epochs &1 \\
\bottomrule
\end{tabular}

        \vspace{-3mm}
        \caption{Outdoor semantic segmentation settings.
        }\label{tab:outdoor_sem_seg_settings}
        \vspace{-4mm}
    \end{minipage}
\end{table}
\nbf{Indoor semantic segmentation.} 
The setting for indoor semantic segmentation is displayed in Tab.~\ref{tab:indoor_sem_seg_settings}. The SP$^2$T model was trained in two phases. In the first stage, emphasis is placed on local fusion, using the local fusion network for separate training on Scannet~\cite{dai2017scannet} or S3DIS~\cite{armeni20163d}. Hence, the model incorporates the weights of local fusion into the second training phase.

\nbf{Indoor instance segmentation.}
Followed by PTv3~\cite{wu2024point}, we use PointGroup~\cite{jiang2020pointgroup} as our foundational framework. Specifically, our configuration mainly follows PTv3. In addition, as with semantic segmentation, the model was trained in two stages.

\begin{table}[!t]
    \begin{minipage}{\linewidth}
    \centering
        \tablestyle{8pt}{1.1}
        \begin{tabular}{lclc}\toprule
\multicolumn{4}{c}{Outdoor Detection}  \\
\cmidrule(lr){1-4} 
Config &Value & Config &Value \\\midrule
optimizer &AdamW  & datasets & Waymo\\
scheduler &Cosine & weight decay &1e-2 \\
framework & CenterPoint & batch size &12 \\
\multicolumn{4}{l}{First Stage:} \\
learning rate &3e-3   & epochs &24\\
block lr scaler & 1e-1 & warmup epochs &0 \\\midrule
\multicolumn{4}{l}{Second Stage:} \\
learning rate &3e-4   & epochs &12\\
block lr scaler & 1.0 & warmup epochs &0 \\
\bottomrule
\end{tabular}

        \vspace{-3mm}
        \caption{Outdoor object detection settings.
        }\label{tab:outdoor_obj_det_settings}
        \vspace{-6mm}
    \end{minipage}
\end{table}

\nbf{Outdoor semantic segmentation.}
Similarly to indoor segmentation, Tab.~\ref{tab:outdoor_sem_seg_settings} outlines the training parameters for SP$^2$T when applied to outdoor segmentation. Similarly to our approach for indoor segmentation, the model undergoes a two-stage training process. In the first stage, a distinct local fusion network is explicitly trained for dataset~\cite{caesar2020nuscenes, behley2019semantickitti}. Then, for the second stage, the model is initialized with the parameters of this local fusion network and continues to train.

\nbf{Outdoor object detection.}
Tab.~\ref{tab:outdoor_obj_det_settings} outlines the training parameters for SP$^2$T when applied to outdoor object detection. The model is also trained through a two-stage approach. Initially, a specific local fusion network is exclusively trained for Waymo~\cite{sun2020scalability}. Subsequently, in the second stage, the model starts with the parameters from this local fusion network, and training is resumed.


\subsection{Model Settings}


Tab.~\ref{tab:appendix_model_settings} presents a comprehensive overview of our model's configuration, focusing primarily on the proxy's initialization and association method,  table-based relative bias and dropout~\cite{srivastava2014dropout}. Furthermore, the parameters for local fusion mirror those of the specific methods~\cite{wu2024point, choy20194d}, and the proxy channel is the same as the channel of local fusion.

\begin{table}[!t]
    \begin{minipage}{\linewidth}
    \centering
        \tablestyle{13pt}{1.1}
        \begin{tabular}{lcc}\\\toprule
Config & Indoor & Outdoor \\\midrule
Proxy embedding depth & \multicolumn{2}{c}{2} \\
Proxy embedding temperature & 10 & 1 \\
Proxy init method & \multicolumn{2}{c}{Spatial-wise} \\
Proxy number & 160 & 400 \\
Proxy search range & [0.0, 1.0] & [0.0, 20.0] \\
Proxy search iter & 10 & 16 \\
Association method & \multicolumn{2}{c}{Vertex-based} \\
Association dim & 3 & 2 \\
Attention channels per head & \multicolumn{2}{c}{16} \\
Attention dropout & \multicolumn{2}{c}{0.0} \\
TRB table size & \multicolumn{2}{c}{16} \\
TRB table strength & \multicolumn{2}{c}{1.0} \\
TRB table temperature & \multicolumn{2}{c}{[0.5, 2.5]} \\
Point-Proxy TRB input scale & 2.5 & 0.2 \\
Proxy-wise TRB input scale & 0.4 & 0.04 \\
Drop path & \multicolumn{2}{c}{0.3} \\
\bottomrule
\end{tabular}
        \vspace{-3mm}
        \caption{Model settings. }\label{tab:appendix_model_settings}
        \vspace{-2mm}
    \end{minipage}
\end{table}

\begin{table}[!t]
    \begin{minipage}{\linewidth}
    \centering
        \tablestyle{1.5pt}{1.1}
        \begin{tabular}{llcc}\toprule
Augmentations &Parameters &Indoor &Outdoor \\\midrule
random dropout &dropout ratio: 0.2, p: 0.2 &\checkmark &- \\
random rotate &axis: z, angle: [-1, 1], p: 0.5 &\checkmark &\checkmark \\
&axis: x, angle: [-1 / 64, 1 / 64], p: 0.5 &\checkmark &- \\
&axis: y, angle: [-1 / 64, 1 / 64], p: 0.5 &\checkmark &- \\
random scale &scale: [0.9, 1.1] &\checkmark &\checkmark \\
random flip &p: 0.5 &\checkmark &\checkmark \\
random jitter &sigma: 0.005, clip: 0.02 &\checkmark &\checkmark \\
elastic distort & params: [[0.2, 0.4], [0.8, 1.6]] &\checkmark &- \\
auto contrast &p: 0.2 &\checkmark &- \\
color jitter &std: 0.05; p: 0.95 &\checkmark &- \\
grid sampling &grid size: 0.02 (indoor), 0.05 (outdoor) &\checkmark &\checkmark \\
sphere crop &ratio: 0.8, max points: 128000 &\checkmark &- \\
normalize color &p: 1 &\checkmark &- \\
\bottomrule
\end{tabular}
        \vspace{-3mm}
        \caption{Data augmentations. }\label{tab:appendix_data_augmentations}
        \vspace{-6mm}
    \end{minipage}
\end{table}

\subsection{Data Augmentations}


As illustrated in Tab.~\ref{tab:appendix_data_augmentations}, we adopted the PTv3~\cite{wu2024point} data augmentation approach to maintain fairness during both training and evaluation. In addition, we applied the same data enhancement to the test set and evaluated other models using this augmentation.

\end{document}